%% file: neurips_2025.tex
\documentclass{article}

\PassOptionsToPackage{numbers, compress}{natbib}

    \usepackage[final]{neurips_2025}

\usepackage[utf8]{inputenc} 
\usepackage[T1]{fontenc}    
\usepackage{hyperref}       
\usepackage{url}            
\usepackage{booktabs}       
\usepackage{amsfonts}       
\usepackage{nicefrac}       
\usepackage{microtype}      
\usepackage{xcolor}         
\usepackage{xpatch}
\usepackage{graphicx}  
\usepackage{threeparttable}  
\usepackage{makecell} 
\usepackage{caption}
\usepackage{xpatch}
\usepackage{microtype}      
\usepackage{tabularx} 
\usepackage{ragged2e} 
\usepackage{rotating}

\usepackage{wrapfig}
\usepackage{mathrsfs}
\usepackage{amssymb}
\usepackage{algorithm}
\usepackage{algorithmic}

\usepackage[algo2e, ruled]{algorithm2e}
\usepackage{multirow}
\usepackage{amsmath}
\usepackage{cleveref}

\makeatletter
\xapptocmd{\NAT@bibsetnum}{\setlength{\leftmargin}{0pt}\setlength{\itemindent}{\labelwidth}\addtolength{\itemindent}{\labelsep}}{}{}
\makeatother

\newcommand{\methodname}[1]{UltraDelta}
\newcommand{\quantizationname}[1]{Uniform Quantization}
\newcommand{\quantizationshort}[1]{UQ}
\newcommand{\pruningname}[1]{Distribution-Aware Compression}
\newcommand{\pruningshort}[1]{DAC}
\newcommand{\allocationname}[1]{Variance-Based Mixed Sparsity Allocation}
\newcommand{\allocationshort}[1]{MSA}
\newcommand{\rescalingname}[1]{Trace-Norm-Guided Rescaling}
\newcommand{\rescalingshort}[1]{TNGR}

\title{Breaking the Compression Ceiling: Data-Free Pipeline for Ultra-Efficient Delta Compression}

\author{
    Xiaohui Wang$^{1}$\thanks{Equal Contribution.} \quad 
    Peng Ye$^{2,4*}$ \quad 
    Chenyu Huang$^{1}$ \quad
    Shenghe Zheng$^{2}$ \quad \\
    \bf{Bo Zhang}$^{2}$ \quad
    \bf{Lei Bai}$^{2}$ \quad
    \bf{Wanli Ouyang}$^{2,4}$ \quad
    \bf{Tao Chen}$^{1,3}$\thanks{Corresponding Author.} \quad \\
    $^1$Fudan University \quad 
    $^2$Shanghai AI Laboratory \quad \\
    $^3$Shanghai Innovation Institute \quad
    $^4$The Chinese University of Hong Kong \\
    \texttt{heatherwang000@gmail.com eetchen@fudan.edu.cn}
}

\begin{document}

\maketitle

\input{paragraphs/0_abstract}
\input{paragraphs/1_introduction}
\input{paragraphs/2_related_work}
\input{paragraphs/3_method}
\input{paragraphs/4_experiment}
\input{paragraphs/5_discussion}

\newpage
\section{Acknowledgement}
This work is supported by National Key Research and Development Program of China (No. 2022ZD0160101), Shanghai Natural Science Foundation (No. 23ZR1402900), Shanghai Science and Technology Commission Explorer Program Project (24TS1401300), Shanghai Municipal Science and Technology Major Project (No.2021SHZDZX0103).
The computations in this research were performed using the CFFF platform of Fudan University.

\bibliographystyle{Ref}  
\small
\bibliography{main}
\normalsize

\newpage
\appendix
\input{paragraphs/appendix}

\end{document}

%% file: paragraphs/0_abstract.tex
\begin{abstract}
With the rise of the fine-tuned–pretrained paradigm, storing numerous fine-tuned models for multi-tasking creates significant storage overhead.
Delta compression alleviates this by storing only the pretrained model and the highly compressed delta weights (the differences between fine-tuned and pretrained model weights). 
However, existing methods fail to maintain both high compression and performance, and often rely on data.
To address these challenges, we propose \methodname{}, the first data-free delta compression pipeline that achieves both ultra-high compression and strong performance. 
\methodname{} is designed to minimize redundancy, maximize information, and stabilize performance across inter-layer, intra-layer, and global dimensions, using three key components:
(1) \allocationname{} assigns sparsity based on variance, giving lower sparsity to high-variance layers to preserve inter-layer information.
(2) \pruningname{} applies uniform quantization and then groups parameters by value, followed by group-wise pruning, to better preserve intra-layer distribution.
(3) \rescalingname{} uses the trace norm of delta weights to estimate a global rescaling factor, improving model stability under higher compression.
Extensive experiments across 
(a) large language models (fine-tuned on LLaMA-2 7B and 13B) with up to 50$\times$ compression, 
(b) general NLP models (RoBERTa-base, T5-base) with up to 224$\times$ compression,
(c) vision models (ViT-B/32, ViT-L/14) with up to 132$\times$ compression, and
(d) multi-modal models (BEiT-3) with 18$\times$ compression, 
demonstrate that \methodname{} consistently outperforms existing methods, especially under ultra-high compression.
Code is available at \href{https://github.com/xiaohuiwang000/UltraDelta}{https://github.com/xiaohuiwang000/UltraDelta}.

\end{abstract}

%% file: paragraphs/1_introduction.tex
\section{Introduction}
As fine-tuning pretrained models for downstream tasks becomes increasingly popular, a growing number of task-specific fine-tuned models have been developed across various domains. 
To obtain multi-task capabilities and achieve optimal performance across tasks, deploying multiple fine-tuned models simultaneously has become a common practice. 
However, multi-model deployment introduces severe storage and computational overhead, since each fine-tuned model requires storing a full set of parameters.
Delta Compression has emerged as a promising solution to this problem by substantially reducing storage requirements. 
Instead of storing multiple complete fine-tuned models, delta compression works by storing a single pretrained model along with a set of delta weights (i.e., the differences between each fine-tuned model and the pretrained model) and then applying aggressive compression to these delta weights to minimize storage overhead.
These delta weights are often highly redundant, allowing for substantial compression with minimal impact on model performance, thereby significantly reducing storage costs.

\begin{figure}
    \centering
    \includegraphics[width=0.98\linewidth]{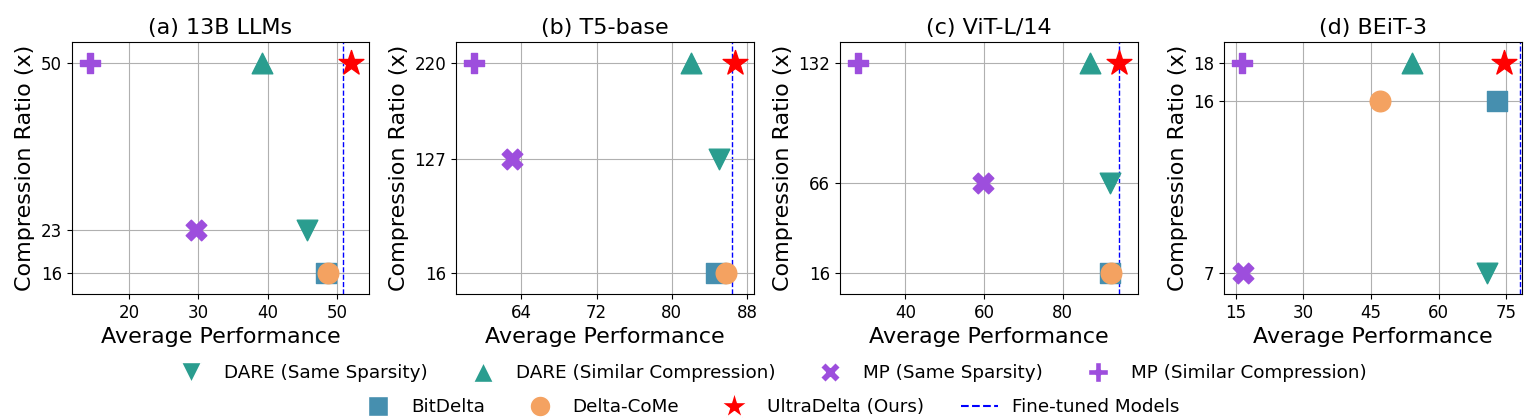}
    \caption{Compression vs. Performance. We compare \methodname{} with baselines, showing strong performance at ultra-high compression. DARE and Magnitude Pruning are evaluated in two settings: with the same sparsity as \methodname{}, and using pruning alone to match a similar compression ratio as \methodname{}. Subfigures show average performance on: (a) LLaMA-2-13B fine-tuned on 3 tasks, (b) T5-base on 8 NLP tasks, (c) ViT-L/14 on 8 vision tasks, and (d) BEiT-3 on 3 vision-language tasks.}
    \label{fig:compression-performance}
    \vspace{-6mm}
\end{figure}

Currently, a number of works have been proposed for delta compression, which can be broadly categorized into two main types: Pruning and Quantization. 
Pruning methods reduce model size by removing parameters. For example, DARE~\cite{yu2024language} randomly prunes delta parameters, while Magnitude Pruning~\cite{han2015deep, han2015learning, li2018optimization, lee2020layer} removes low-magnitude parameters.
Quantization methods work by mapping full-precision parameters to low-bit representations. For example, BitDelta~\cite{liu2024bitdelta} uses masking and rescaling to achieve 1-bit quantization, while Delta-CoMe~\cite{ping2024delta} applies mixed-precision quantization to matrices obtained from singular value decomposition.
Other approaches, such as DeltaZip~\cite{yao2025deltazip}, combine quantization with pruning to improve deployment efficiency.
However, these methods all struggle to balance ultra-high compression with strong performance, mainly due to: 
(1) Ignoring inter-layer differences: All layers are treated equally, ignoring that different layers contribute unequally to the model, which limits information preservation.
(2) Disrupting intra-layer distributions: Intra-layer weight distributions are crucial to performance but are often distorted by aggressive quantization or pruning.
(3) Lack of stability under ultra-high compression: Existing methods struggle to maintain stability under ultra-high compression without relying on data, leading to severe performance drops.

To address these limitations, we propose \methodname{}, the first data-free pipeline to achieve ultra-efficient delta compression, delivering both ultra-high compression and strong performance.
\methodname{} focuses on minimizing parameter redundancy, maximizing information preservation, and enhancing model stability across three dimensions: inter-layer, intra-layer and global. It consists of three key components:
(1) \allocationname{} (\allocationshort{}, inter-layer):
We theoretically show that layer-wise variance reflects the amount of information in each layer. Based on this insight, we assign lower sparsity to layers with higher variance to better preserve critical information.
(2) \pruningname{} (\pruningshort{}, intra-layer):
We first apply uniform quantization, and group parameters by their quantized value, then perform random pruning within each group to maintain the relative proportions across different values and better preserve the original distribution.
(3) \rescalingname{} (\rescalingshort{}, global):
We observe that under extreme sparsity, the standard rescaling factor $1/(1-s)$ becomes insufficient to stabilize performance (where $s$ is the sparsity rate). We introduce a refined rescaling factor $\gamma/(1-s)$, where $\gamma$ is heuristically estimated from the trace norm of each delta weight and is smaller for delta weights with larger trace norms, enhancing robustness under ultra-high compression.

We conduct extensive experiments across models of different scales, types, and tasks to evaluate the effectiveness and robustness of \methodname{}. The results demonstrate its exceptional performance under ultra-high compression:
(1) Large Language Models: \methodname{} compresses LLaMA-2 series models~\cite{touvron2023llama,luo2023wizardmath,luo2023wizardcoder} to 32$\times$ compression for 7B and 50$\times$ for 13B, consistently outperforming all baselines and even surpassing the average performance of fine-tuned models. To further demonstrate generality, we also evaluate on more recent architectures, including LLaMA-3.1~\cite{lambert2024tulu3}, Qwen2.5~\cite{qwen2}, and Qwen3~\cite{qwen3guard}.
(2)	General NLP Models: \methodname{} achieves 224$\times$ compression on T5-base~\cite{raffel2020exploring} and 32$\times$ on RoBERTa-base~\cite{liu2019roberta} across 8 NLP tasks, even exceeding the average performance of fine-tuned models on T5-base.
(3)	Vision Models: \methodname{} achieves 50$\times$ compression on ViT-B/32 and 132$\times$ on ViT-L/14~\cite{radford2021learning} across 8 image classification tasks, achieving completely lossless performance on ViT-L/14 compared with fine-tuned models.
(4)	Multi-modal Models: On BEiT-3~\cite{beit3}, we achieve 18$\times$ compression on 3 vision-language tasks, outperforming all baselines across all tasks.
As illustrated in Fig.~\ref{fig:compression-performance}, we present compression–performance trade-off for all four model types. \methodname{} consistently achieves the best trade-off, outperforming all baselines, and in some cases, even surpassing the performance of fine-tuned models under ultra-high compression.

The main contributions of this paper can be summarized as follows:

\noindent $\bullet$ To break the compression ceiling of delta weights, we analyze the limitations of existing methods in information preservation and model stability, and propose \methodname{}, the first data-free pipeline enabling ultra-efficient delta compression, achieving both ultra-high compression ratios and strong performance without relying on any data.

\noindent $\bullet$ To enhance information preservation and model stability, we introduce three novel techniques: \allocationname{} to prioritize critical layers, \pruningname{} to preserve the original weight distribution, and \rescalingname{} to stabilize model performance under extremely sparsity.

\noindent $\bullet$ Extensive experiments across large language models, general NLP models, vision models, and multi-modal models demonstrate that \methodname{} consistently outperforms all baselines and even surpasses fine-tuned models in several settings, showcasing its effectiveness and robustness in achieving ultra-efficient delta compression.

%% file: paragraphs/2_related_work.tex
\section{Related Work}
Delta Compression is a technique that compresses the delta weights. 
It is especially beneficial in multi-model deployment scenarios, where many task-specific models are fine-tuned from the same pretrained model. 
By storing only the highly compressed delta weights alongside the shared pretrained model, delta compression significantly eliminates redundant storage. 
Existing methods for delta compression primarily fall into two categories: pruning and quantization.

\textbf{Delta Weight Pruning} reduces model size by removing parameters from the delta weights.
Magnitude Pruning (MP)\cite{han2015deep,han2015learning,lee2020layer,li2018optimization} removes weights with the smallest magnitudes but fails to preserve information and causes significant performance drops under high sparsity, as it disrupts the original weight distribution. 
DARE\cite{yu2024language} applies random pruning combined with global rescaling based on sparsity to improve robustness, but suffers from instability under high sparsity.
DAREx~\cite{deng2024dare} refines DARE’s rescaling using activation statistics, and DeltaDQ~\cite{jiang2024deltadq} performs group-wise pruning and selects the optimal group size based on attention error. However, both methods require data for tuning, limiting their use in data-free scenarios.
Some methods also adopt layer-wise pruning~\cite{li2024adaptive, lu2024alphapruning}, but the metrics are often computationally expensive and usually require data.
Overall, pruning effectively reduces parameter redundancy but suffers from several drawbacks, such as insufficient information preservation, instability under high sparsity, and reliance on additional data.

\textbf{Delta Weight Quantization} compresses delta weights by mapping them to low-bit representations.
GPT-Zip\cite{isik2023gpt} extends GPTQ\cite{frantar2022gptq} and quantizes delta parameters to 2-bit. 
Delta-DCT~\cite{huang2025seeing} leverages the discrete cosine transform and compresses delta weights in the frequency domain.
BitDelta~\cite{liu2024bitdelta} uses binary masks and scaling factors for 1-bit quantization, but requires calibration data and also distorts the original weight distribution due to the binary masks.
Delta-CoMe~\cite{ping2024delta} applies singular value decomposition and mixed-precision quantization on the decomposed matrices to achieve 1-bit compression.
Overall, although quantization effectively reduces parameter bit-width, it is often limited to 1-bit precision, which prevents further minimization of parameter redundancy.


\textbf{Hybrid Approach} combines delta weight pruning and delta weight quantization to leverage the advantages of both.
DeltaZip~\cite{yao2025deltazip} employs structured pruning and quantization, primarily aiming to improve deployment efficiency and hardware acceleration, but it falls short in preserving critical information, limiting its effectiveness under ultra-high compression.
ComPEFT~\cite{yadav2023compeft}, on the other hand, applies magnitude pruning and then quantizes the remaining weights with a fixed shared value. However, it requires validation data and, due to the reliance on magnitude pruning, disrupts structural integrity and fails to preserve critical information under high sparsity.

In contrast, \methodname{} is a data-free hybrid compression method that explicitly focuses on minimizing parameter redundancy, maximizing information preservation, and enhancing model stability under ultra-high compression ratios, all without requiring any data.

%% file: paragraphs/3_method.tex
\section{Method}

\begin{figure}
    \centering
    \includegraphics[width=\linewidth]{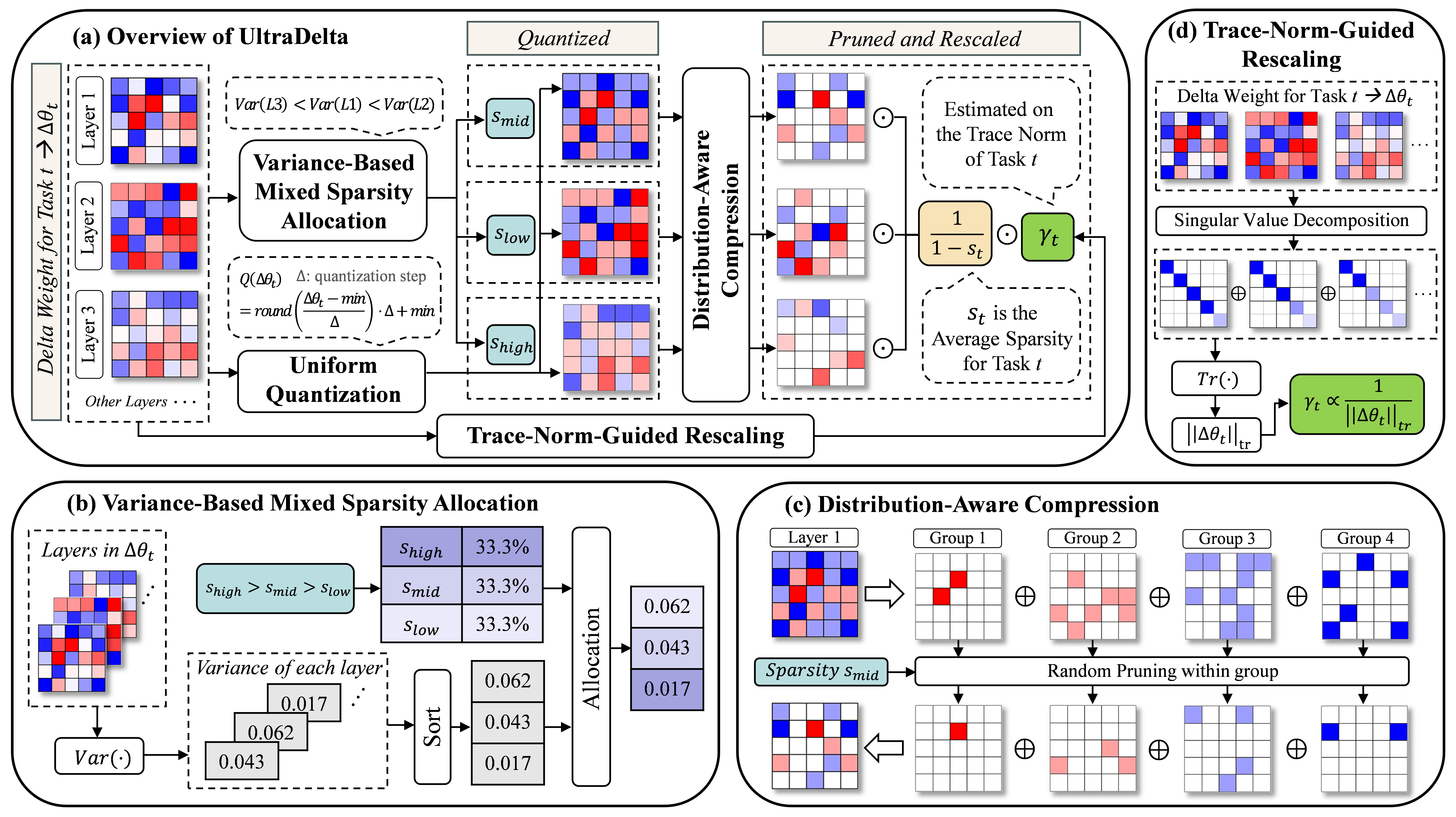}
    \caption{(a) Overview of the \methodname{} pipeline. It first assigns layer-wise sparsity based on variance, and performs uniform quantization followed by group-wise pruning, and finally applies trace-norm-guided rescaling. (b) \allocationname{} allocates lower sparsity to high-variance layers to preserve information. (c) \pruningname{} groups quantized parameters by value and prunes within each group to retain distribution. (d) \rescalingname{} estimates a rescaling factor using the trace norm of each delta weight to enhance robustness.}
    \label{fig:method}
    \vspace{-2mm}
\end{figure}

\subsection{Preliminaries}
Consider a set of $T$ fine-tuned models that are fine-tuned from the same pretrained model. Let the weights of these models be denoted as $\{\theta_1,\theta_2,\ldots,\theta_T\}$, and the shared pretrained model weight as $\theta_{pre}$. For each fine-tuned model $t\in\{1,2,\ldots,T\}$, we define the delta weight as:
\begin{equation}
    \Delta\theta_t=\theta_t-\theta_\mathrm{pre}
\end{equation}

\subsection{\methodname{}: A Data-Free Pipeline for Ultra-Efficient Delta Compression}
\label{method:components}

To address the limitations of existing delta compression methods, \methodname{} introduces a hybrid compression framework that minimizes parameter redundancy while maximizing information preservation and enhancing model stability across three dimensions: inter-layer, intra-layer, and global, enabling robustness under ultra-high compression.
An overview of the overall pipeline is illustrated in Fig.~\ref{fig:method}.
We detail the core design of \methodname{} in Sec.\ref{method:components}, and provide analysis in Sec.\ref{method:theory}.

\textbf{(1) \allocationname{} (\allocationshort{}).}
Existing pruning methods often adopt uniform sparsity across layers, treating all parts of the model equally and overlooking their varying importance. However, different layers exhibit different sensitivities to pruning.
We observe that layers with higher variance tend to carry more critical information and are more sensitive to pruning (see Sec.~\ref{method:theory}).
Motivated by this, we propose \allocationshort{} that categorizes all layers into three groups based on their variances (low, medium, high), each containing an equal number of parameters:
\begin{equation}\mathrm{Groups}=\{\mathcal{G}_{\text{low}},\mathcal{G}_{\text{mid}},\mathcal{G}_{\text{high}}\}\end{equation}
Each group $\mathcal{G}_{v}$ is assigned a sparsity rate $s_{v}$ based on its variance $v \in \{\text{low}, \text{mid}, \text{high}\}$, with lower-variance groups assigned higher sparsity and higher-variance groups assigned lower sparsity, in order to preserve inter-layer information better:
\begin{equation}
    s_{v} = s_{\text{mid}} + \delta_v, \quad \delta_v \in \left\{ +s_{\text{step}}, 0, -s_{\text{step}} \right\}
    \label{equation:step_size}
\end{equation}
where $s_{\text{mid}}$ is the average sparsity of the overall delta weight, and $s_{\text{step}} > 0$ controls the difference in sparsity across groups.

\textbf{(2) \pruningname{} (\pruningshort{}).}
Prior work has shown that preserving the shape of the weight distribution after compression helps maintain model performance~\cite{huang2025seeing}. 
To better preserve the intra-layer distribution shape under ultra-high compression, we propose \pruningshort{}. 
Specifically, as delta weight exhibits a distribution well-suited to uniform quantization~\cite{jiang2024deltadq}, we first apply uniform quantization with a lower bit width $b$ (typically $b=4$) to the delta weight. 
This quantization process consists of two steps: (a) mapping each value in $\Delta\theta_t$ to a discrete integer within $[0,2^b-1]$; (b) mapping the quantized values back to the original value range. Let $\min = \min(\Delta\theta_t)$, $\max = \max(\Delta\theta_t)$, and $\Delta$ be the quantization step size:
\begin{equation}
\label{equation:quantization}
\hat{\Delta\theta}_t=\mathrm{round}\left(\frac{\Delta\theta_t-\min}{\Delta}\right)\cdot\Delta+\min,\quad\mathrm{where}\quad\Delta=\frac{\max-\min}{2^b-1}
\end{equation}
After quantization, we group the parameters by their values and perform random pruning within each group.
Let $\{u_1, u_2, \dots, u_n\}$ represent the unique values in the quantized delta weight $\hat{\Delta\theta_t}$. We denote $(j,k)$ as the row and column indices of elements in $\hat{\Delta\theta_t}$.
For each unique quantized value $u_i$, we define a mask $\boldsymbol{M}_{u_i}$ whose elements are independently sampled from a Bernoulli distribution with success probability $1 - s_l$, where $s_l$ is the sparsity rate of the layer. Specifically:
\begin{equation}
\boldsymbol{M}_{u_i}^{(j,k)} \sim 
\begin{cases}
\mathrm{Bernoulli}(1 - s_l), & \text{if } \hat{\Delta\theta_t}^{(j,k)} = u_i \\
0, & \text{otherwise}
\end{cases}
\label{eq:mask_definition1}
\end{equation}
The pruned delta weight $\hat{\Delta\theta_t}^*$ is obtained by applying the group-wise Bernoulli masks to the quantized delta weights:
\begin{equation}
\hat{\Delta\theta_t}^* = \sum_{i=1}^n u_i\cdot \boldsymbol{M}_{u_i}
\end{equation}
\pruningshort{} preserves the relative proportions of delta values, maintaining the intra-layer distribution shape. It minimizes distortion under high sparsity and ensures more stable performance.
Moreover, \pruningshort{} can be applied in non-quantized settings with minimal modification, making it suitable for scenarios where aggressive compression is not required. Instead of grouping by discrete values, we partition the weight range into several intervals, and assign parameters whose values fall within the same interval to the same group. Details are provided in App.~\ref{appendix:noquant_das}.

\textbf{(3) \rescalingname{} (\rescalingshort{}).}
As derived in Sec.~\ref{method:theory}, the variance of activation errors between compressed and original delta weights correlates with $s/(1-s)$, causing instability at high sparsity.
To address this, we introduce an additional rescaling factor $\gamma_t$ for task $t$, and redefine rescaling as $\gamma_t/(1-s_t)$, where $s_t$ is the overall sparsity of the delta weight for task $t$.
We find that delta weights with larger trace norms (i.e., the sum of their singular values) tend to require smaller $\gamma_t$ to maintain stable performance. Motivated by this observation, we set $\gamma_t$ inversely proportional to the trace norm of the delta weights:
\begin{equation}
\gamma_t \propto \frac{1}{||\Delta\theta_t||_{\mathrm{tr}}}
\end{equation}
where $\|\cdot\|_{\mathrm{tr}}$ denotes the trace norm. This trace-norm-guided rescaling offers a simple yet effective way to adaptively stabilize pruning under extreme sparsity. In practice, $\gamma_t$ typically falls within the range of $[0.5,1.0]$.
During the inference stage, the final model weight is reconstructed as:
\begin{equation}
\theta_t^{\mathrm{final}} = \theta_{\mathrm{pre}} + \frac{\gamma_t}{1 - s_t} \cdot \hat{\Delta\theta_t}^*
\end{equation}

\subsection{Reason for Effectiveness}
\label{method:theory}
\subsubsection{Theoretical Analysis of \allocationshort{}}
\textbf{In lossless compression,} the entropy represents the lower bound on the achievable average bit rate. 
A larger entropy means that a layer carries more information and therefore should be more carefully preserved. 
We show that the variance of a layer is closely related to its entropy, serving as the theoretical motivation for \allocationshort{}.
Following~\cite{lin2024mlp}, we model the distribution of a layer as a Gaussian Distribution $L\sim{\mathcal{N}}(\mu,\sigma^2)$. The probability density function of $L$ is given by:
\begin{equation}p(l)=\frac{1}{\sqrt{2\pi\sigma^2}}\exp\left(-\frac{(l-\mu)^2}{2\sigma^2}\right)\end{equation}
The entropy of this distribution is (detailed derivation in App.~\ref{appendix:variance_entropy}):
\begin{equation}H(L)=-\mathbb{E}[\log p(l)]=\log(\sigma)+\frac{1}{2}\log(2\pi)+\frac{1}{2}\end{equation}
This demonstrates the link between variance and entropy: layers with larger variance have higher entropy, implying greater information content and thus requiring smaller sparsity to preserve it.

\textbf{In lossy compression,} entropy alone is not sufficient. 
When distortion is allowed, we employ rate–distortion theory~\cite{berger2003rate,cover1999elements}, which establishes the limit on how much information must be retained to achieve a given distortion level.
Let $\hat{L}$ denote the compressed version of $L$. 
With mean squared error (MSE) distortion, we define:
\begin{equation}
D = \mathbb{E}\big[(L-\hat{L})^2\big],
\end{equation}
According to rate–distortion theory, the rate–distortion function of a Gaussian source under MSE distortion (which is independent of the mean $\mu$) is given by:
\begin{equation}
R(D)=
\begin{cases}
\frac{1}{2}\log\left(\frac{\sigma^2}{D}\right), & 0<D<\sigma^2, \\
0, & D\geq\sigma^2, 
\end{cases}
\end{equation} 
This shows that, for a fixed distortion $D$, a larger variance $\sigma^2$ leads to a higher rate $R(D)$, meaning that more information must be retained. 
In other words, layers with larger variance require a smaller sparsity in order to preserve their information under lossy compression.


\begin{figure}
    \centering
    \includegraphics[width=0.97\linewidth]{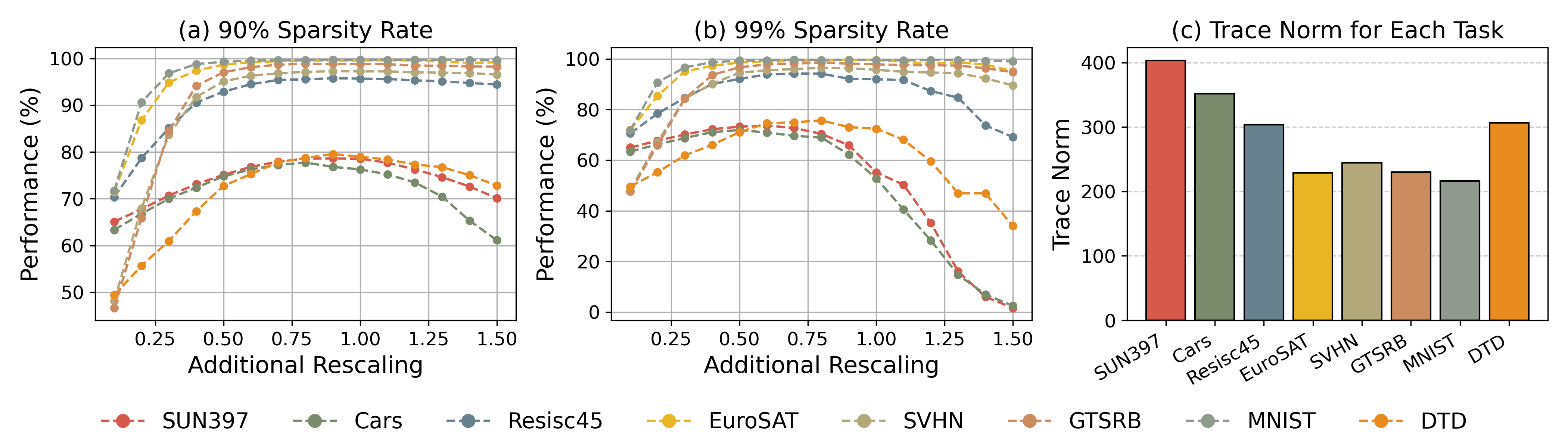}
    \caption{Relationship between Trace Norm and Additional Rescaling Factor on 8 ViT-B/32.
    (a) Average performance under 90\% sparsity with different additional rescaling factors.  
    (b) Same as (a), under 99\% sparsity.  
    (c) Trace norm of delta weights across tasks.}
    \label{fig:trace-norm}
    \vspace{-3mm}
\end{figure}

\subsubsection{Theoretical Analysis of the Rescaling Factor.}
For a given delta weight $\Delta\theta$, let the activation be defined as $a=\Delta\theta\odot x$, where $x$ is the input feature. We introduce a Bernoulli mask $B\sim\mathrm{Bernoulli}(1-s)$, with sparsity rate $s$, and apply an additional scaling factor $\gamma\in[0,1]$. Following~\cite{deng2024dare}, the activation error introduced by compression is:
\begin{equation}
\label{equation:activation_error}
\varepsilon=\Delta\theta\odot x-\frac{\gamma}{1-s}\cdot(B\odot\Delta\theta\odot x)=a\odot\left(1-B\cdot\frac{\gamma}{1-s}\right)
\end{equation}
We further derive the variance of this error (detailed derivation in App.~\ref{appendix:activation_error}):
\begin{equation}\mathrm{Var}(\varepsilon)=\frac{\gamma^2s}{1-s}\odot a^2\end{equation}
As sparsity becomes higher, the variance grows rapidly and causes instability. This underscores the importance of using smaller rescaling factors to stabilize the model under high sparsity.

\subsubsection{Empirical Analysis of \rescalingshort{}.}
We investigate how the additional rescaling factor $\gamma$ for each delta weight correlates with its intrinsic characteristics, as shown in Fig.~\ref{fig:trace-norm}. 
At 90\% sparsity, we observe that delta weights with larger trace norms tend to require smaller values of $\gamma$ to maintain stable performance. Moreover, their performance is more sensitive to changes in $\gamma$.
This phenomenon becomes even more pronounced at 99\% sparsity, where the need for smaller $\gamma$ is more substantial for delta weights with large trace norms, and the performance drop-off becomes steeper when $\gamma$ is not well matched.
These empirical trends support our trace-norm-guided heuristic estimation, which sets $\gamma \in [0.5, 1.0]$ inversely proportional to the trace norm for adaptive and data-free rescaling.

\begin{wrapfigure}{r}{0.6\textwidth}
\vspace{-5mm}
  \centering
  \includegraphics[width=0.6\textwidth]{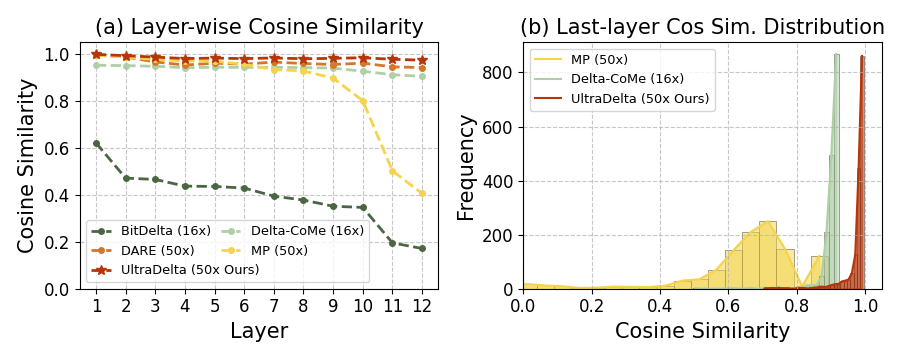}
  \caption{Cosine Similarity Analysis. (a) Layer-wise similarity of each layer's embeddings. (b) Distribution of cosine similarity for the final-layer embeddings. Compression ratios for each method are indicated in parentheses.}
  \label{fig:cosine_similarity}
  \vspace{-3mm}
\end{wrapfigure}

\subsubsection{Ability to Preserve Information.}
To explain why \methodname{} achieves strong performance under ultra-high compression ratios, we analyze how well it preserves the original model's internal representations. Specifically, we compare the cosine similarity between the embeddings from the compressed and fine-tuned RoBERTa-base~\cite{liu2019roberta} models on the CoLA~\cite{warstadt2019neural} dataset. We extract the layer-wise embeddings of each input token and report the average cosine similarities. As shown in Fig.~\ref{fig:cosine_similarity}, our method consistently maintains high similarity across all layers, especially in the deeper ones that are more critical to performance. The distribution of cosine similarities in the final layer further confirms that \methodname{} retains essential information, indicating its effectiveness in preserving critical information despite ultra-high compression.

%% file: paragraphs/4_experiment.tex
\section{Experiment}
\textbf{Baseline Methods.} 
We compare \methodname{} with the following baselines: 
(1) Fine-tuned Models; 
(2) Multi-task Learning (MTL)~\cite{caruana1997multitask}, which trains a single model on all tasks; 
(3) BitDelta~\cite{liu2024bitdelta}; 
(4) Delta-CoMe~\cite{ping2024delta}; 
(5) DARE~\cite{yu2024language} with either the same sparsity or similar compression ratio as \methodname{}, using pruning alone to achieve a similar compression ratio; 
(6) Magnitude Pruning (MP)~\cite{han2015deep}, also matched by sparsity or compression ratio.
Since BitDelta and Delta-CoMe target only linear layers in Transformer blocks, we restrict all methods to compress only linear layers for fairness. All original models are stored in FP16 format. Experiments are conducted on a NVIDIA A800 GPU.

\textbf{Compression Ratio.}
Unlike quantization, where compression is directly determined by bit-width, pruning methods require storing the indices of non-zero values.
To ensure a unified and fair evaluation, following~\cite{sattler2019sparse,strom15_interspeech,yadav2023compeft}, we adopt Golomb coding~\cite{golomb1966run} to store the compressed delta weights for methods involving pruning, which is well-suited for encoding the zero-run lengths that approximately follow a geometric distribution. Detailed derivation and calculations are in App.~\ref{appendix:golomb}, with results summarized in Tab.~\ref{tab:perf_all}.
Furthermore, for a complete evaluation, we present the ideal upper bound of compression ratio and practical results under different storage schemes in Sec.~\ref{ablation_study}(storage scheme).

\subsection{Performance on Large Language Models (LLMs)}
\textbf{Settings.} 
We primarily evaluate \methodname{} on the LLaMA-2 series with sizes of 7B and 13B across three types of models: math (WizardMath~\cite{luo2023wizardmath}), code (WizardCoder~\cite{luo2023wizardcoder}), and chat (LLAMA-2-Chat~\cite{touvron2023llama}). 
The models are evaluated using GSM8K~\cite{cobbe2021training} for math (accuracy), HumanEval~\cite{chen2021evaluating} for code (pass@1), and TruthfulQA~\cite{lin2021truthfulqa} for chat (accuracy).
We also incorporate more recent LLMs and more challenging tasks: LLaMA-3.1-Tulu-8B~\cite{lambert2024tulu3} evaluated on MBPP+~\cite{austin2021program} and HumanEval+~\cite{chen2021evaluating}, Qwen2.5-7B-Instruct~\cite{qwen2} evaluated on MATH~\cite{hendrycks2021measuring} and GPQA~\cite{rein2024gpqa}, and Qwen3Guard-8B~\cite{qwen3guard} evaluated on MMLU~\cite{hendryckstest2021} and BBQ~\cite{parrish2021bbq}.
Details for the LLM IDs are presented in App.~\ref{appendix:model_ids}.

\textbf{Results.} 
We first report results on the LLaMA-2 series (see Tab.~\ref{tab:performance_llm}).
For the 7B model, we apply 4-bit quantization and prune 95\% of parameters, achieving an 32.9$\times$ compression ratio; for the 13B model, we prune 97\% to reach 50.9$\times$ compression.
Despite these ultra-high compression ratios, \methodname{} achieves average scores of 45.57 (7B) and 52.05 (13B), exceeding the fine-tuned models (45.37 and 50.94, respectively). This improvement suggests that \methodname{} may introduce a regularization benefit.
Compared to BitDelta~\cite{liu2024bitdelta} and Delta-CoMe~\cite{ping2024delta}, which are limited to 16$\times$ compression, \methodname{} delivers both higher compression and better performance. 
Other approaches such as DARE~\cite{yu2024language} and MP~\cite{han2015deep} fall short even at lower compression ratios, and when matched to the similar compression ratio as \methodname{}, their performance drops sharply.
For newer LLMs (see Tab.~\ref{tab:performance_recent_llm}), \methodname{} achieves the best overall compression–performance trade-off across baselines.
In summary, \methodname{} consistently outperforms all baselines across various tasks and model sizes, demonstrating its remarkable robustness and effectiveness under ultra-high compression.

\input{tables/3_llm}
\input{tables/3_recent_llm}

\subsection{Performance on general NLP models}
\textbf{Settings.} Following~\cite{tang2024fusionbench,yu2024language}, we evaluate both T5-base~\cite{raffel2020exploring} and RoBERTa-base~\cite{liu2019roberta} models on the GLUE~\cite{wang2018glue} benchmark, covering CoLA~\cite{warstadt2019neural}, SST-2~\cite{socher2013recursive}, MRPC~\cite{dolan2005automatically}, STS-B~\cite{cer2017semeval}, QQP~\cite{iyer2017first}, MNLI~\cite{williams2017broad}, QNLI~\cite{rajpurkar2016squad}, and RTE~\cite{giampiccolo2007third}. For T5-base, we report Spearman’s $\rho$ on STS-B and accuracy on the other tasks. Settings and results of RoBERTa-base are provided in App.~\ref{appendix:roberta_results}.

\textbf{Results.} 
The results for T5-base models are presented in Tab.~\ref{tab:performance_t5}. 
Using 4-bit quantization combined with 99.5\% pruning, \methodname{} achieves an impressive 224.6$\times$ compression ratio.
Despite this extreme compression, \methodname{} attains an average accuracy of 86.74, outperforming all existing compression baselines and even slightly exceeding the full fine-tuned model’s accuracy of 86.37.
Compared to BitDelta~\cite{liu2024bitdelta} and Delta-CoMe~\cite{ping2024delta}, \methodname{} achieves a much higher compression ratio while still maintaining superior performance. This demonstrates that \methodname{} successfully pushes the compression limits of delta weights while preserving model performance, highlighting its effectiveness and strong generalization ability.

\input{tables/8_t5}

\subsection{Performance on Vision Models}
\textbf{Settings.} Following~\cite{ilharco2022editing}, we evaluate ViT-B/32 and ViT-L/14~\cite{radford2021learning} models on eight image classification datasets: SUN397~\cite{xiao2010sun}, Stanford Cars~\cite{krause20133d}, RESISC45~\cite{cheng2017remote}, EuroSAT~\cite{helber2019eurosat}, SVHN~\cite{netzer2011reading}, GTSRB~\cite{stallkamp2011german}, MNIST~\cite{lecun1998mnist}, and DTD~\cite{cimpoi2014describing}. 
Accuracy is used as the evaluation metric for all datasets. 

\textbf{Results.}
The results of ViT-L/14 models are shown in Tab.~\ref{tab:performance_vitlarge14}, while ViT-B/32 results are provided in App.~\ref{appendix:vitb_results}.
With 4-bit quantization and 99\% sparsity, \methodname{} achieves a 132.5$\times$ compression ratio while maintaining an average accuracy of 94.4, matching the full fine-tuned model and significantly outperforming all compressed baselines, demonstrating its effectiveness on vision models.

\input{tables/8_vit_l_14}

\subsection{Performance on multi-modal models}
\textbf{Settings.}
We compress delta weights on BEiT-3~\cite{beit3} models fine-tuned on three datasets: VQA~\cite{goyal2017making} (Visual Question Answering), NLVR2~\cite{suhr2018corpus} (Visual Reasoning), and COCO Captioning~\cite{lin2014microsoft} (Image Captioning). The COCO Captioning task is evaluated using BLEU4~\cite{papineni2002bleu}, CIDEr~\cite{vedantam2015cider}, METEOR~\cite{banerjee2005meteor}, and ROUGE-L~\cite{lin2004rouge}, while the other two tasks are evaluated based on accuracy.

\textbf{Results.} 
The results of BEiT-3 models are shown in Tab.~\ref{tab:performance_beit}. 
With 4-bit quantization and 90\% sparsity, \methodname{} achieves a 18.4$\times$ compression ratio maintaining high accuracy and quality, outperforming all baselines across all tasks, demonstrating its effectiveness on multi-modal models.

\input{tables/3_beit3}
\input{tables/abl_component}

\subsection{Ablation Study}
\label{ablation_study}
\begin{wrapfigure}{r}{0.35\textwidth}
\vspace{-5mm}
  \centering
  \includegraphics[width=0.35\textwidth]{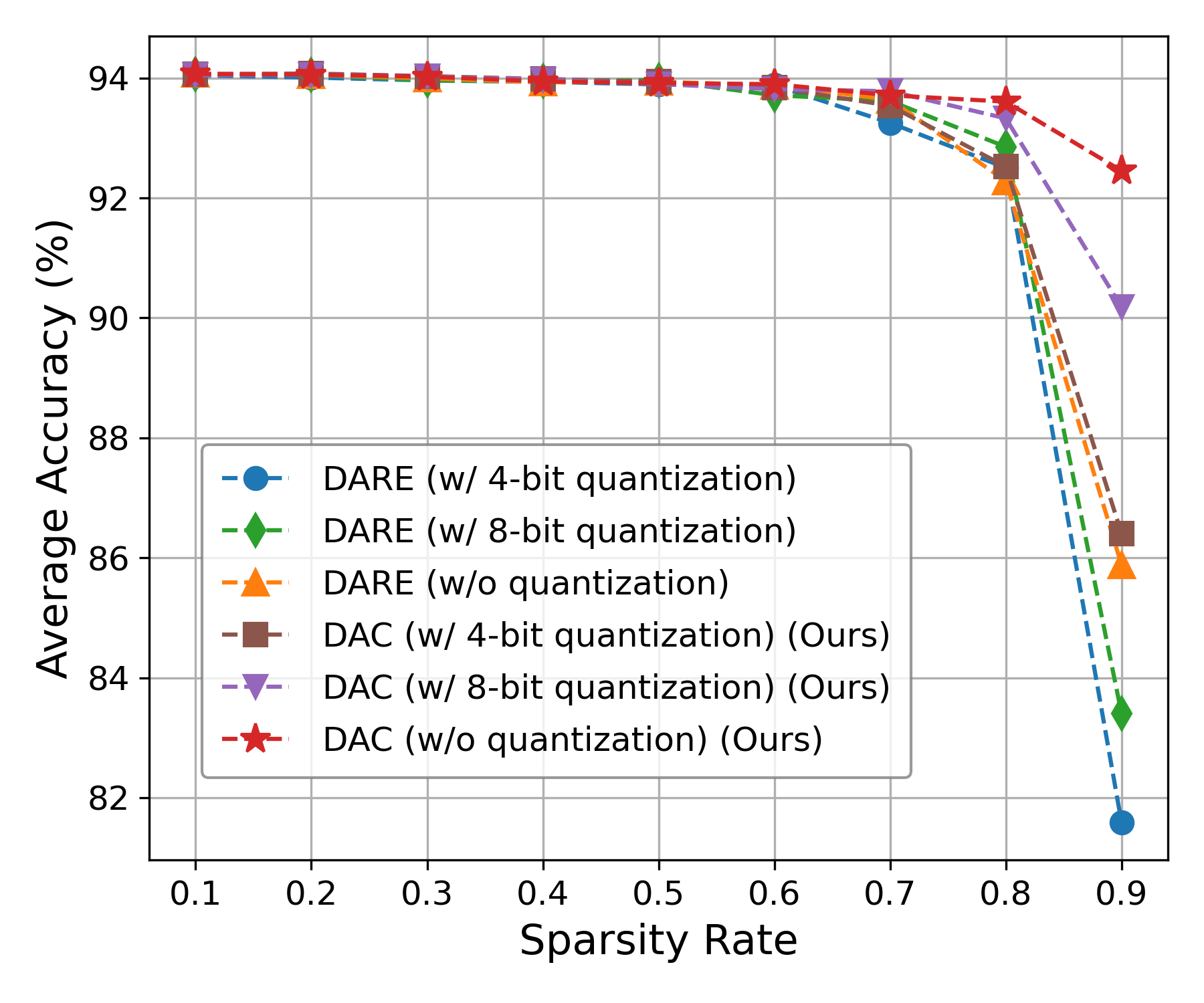}
  \caption{Average Performance of \pruningshort{} and DARE on 30 ViT-B/16.}
  \label{fig:ablation_30vit}
  \vspace{-5.5mm}
\end{wrapfigure}

\textbf{Effectiveness of Each Component.}
We evaluate the contribution of each component on 8 ViT-B/32 in Tab.~\ref{tab:ablation_component}. \pruningshort{} substantially improves compression ratio from 23.7$\times$ to 50.9$\times$, all while preserving model accuracy. \allocationshort{} yields the largest accuracy improvement, highlighting the effectiveness of mixed sparsity allocation. 
Together, they enable ultra-high compression with strong performance.

\textbf{Effectiveness of \pruningshort{}.}
To further validate the robustness of \pruningshort{}, we evaluate it on a large-scale benchmark~\cite{huang2024emr} of 30 ViT-B/16 models across 4-bit, 8-bit, and non-quantized configurations. For fair comparison, we apply uniform quantization to both methods in quantized cases. As shown in Fig.~\ref{fig:ablation_30vit} (detailed results are in App.~\ref{appendix:ablation_dac}), \pruningshort{} consistently outperforms DARE~\cite{yu2024language} across all configurations.
Notably, under 90\% sparsity and 4-bit quantization, \pruningshort{} still exceeds DARE’s non-quantized performance, confirming its robustness across diverse tasks.

\begin{wrapfigure}{r}{0.60\textwidth}
\vspace{-4mm}
  \centering
  \includegraphics[width=0.60\textwidth]{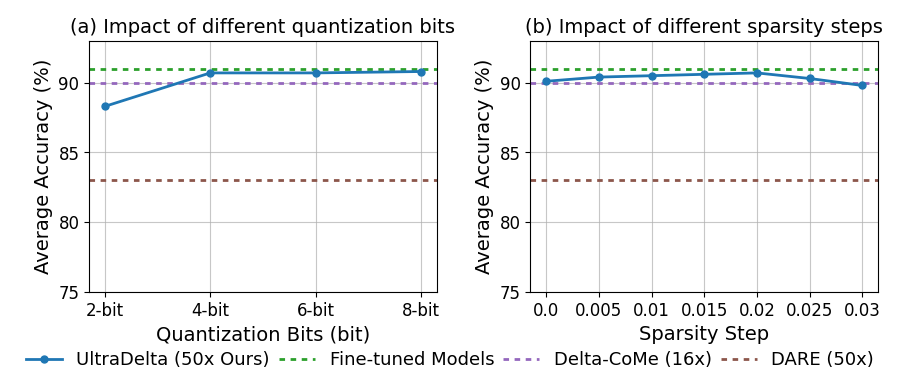}
  \caption{Impact of hyper-parameter on 8 ViT-B/32. (a) Impact of quantization bits; (b) Impact of sparsity steps.}
  \label{fig:ablation_hyper}
  \vspace{-3mm}
\end{wrapfigure}

\textbf{Impact of Hyper-Parameters.} We evaluate key hyper-parameters in \pruningshort{} and \allocationshort{} on 8 ViT-B/32 models, as shown in Fig.~\ref{fig:ablation_hyper} (detailed results are in App.~\ref{appendix:ablation_hyper}).
For bit-width in quantization (see Eq.~\ref{equation:quantization}), accuracy improves notably from 2-bit to 4-bit, with marginal gains beyond, suggesting 4-bit as an optimal balance.
For sparsity step $s_\text{step}$ in \allocationshort{} (see Eq.~\ref{equation:step_size}), we evaluate it under a fixed 97\% target sparsity and 4-bit quantization. 
A moderate step size (within $[0.01,0.02]$) yields the best average performance. 
Performance remains stable as long as extremely large or small step sizes are avoided. This indicates that precise tuning of $s_\text{step}$ is generally unnecessary.

\textbf{Storage Scheme.} 
We report compression ratios of \methodname{} under different storage schemes. 
Index-Free Storage denotes the ideal upper bound where only non-zero values are stored without indices (the ideal compression ratios of other pruning baselines are in Tab.~\ref{tab:perf_all}). 
We also evaluate practical schemes, including Golomb coding, Compressed Sparse Row (CSR), and Block Compressed Sparse Row (BCSR). 
As shown in Tab.~\ref{tab:storage_formats_transposed}, BCSR performs poorly, as it only works when zeros are densely clustered and thus is not suitable for our setting.

\input{tables/abl_storage}

%% file: tables/3_llm.tex
\begin{table*}[htbp]

\centering
\caption{Performance comparison on large language models across 3 tasks. Ratio denotes the compression ratio (original model size / compressed size). Methods marked with ``(Sparsity)'' use the same sparsity rate as ours, while those marked with ``(Compression)'' use a higher sparsity rate to match a similar overall compression ratio as our method. Best results are highlighted in bold.}
\label{tab:performance_llm} 
\resizebox{0.99\textwidth}{!}{
\begin{tabular}{l|cc|cc|cc|cc|cc}
\toprule

\multirow{2}{*}{\textbf{Method}} 
& \multicolumn{2}{c|}{\textbf{Ratio}} 
& \multicolumn{2}{c|}{\textbf{WizardMath-V1.0}} 
& \multicolumn{2}{c|}{\textbf{WizardCoder-V1.0}} 
& \multicolumn{2}{c|}{\textbf{LLAMA-2-Chat}} 
& \multicolumn{2}{c}{\textbf{Avg}} \\
& 7B & 13B & 7B & 13B & 7B & 13B & 7B & 13B & 7B & 13B \\
\midrule

\multicolumn{1}{l|}{Individual} & 1$\times$ & 1$\times$ & 41.55 & 47.31 & 49.40 & 58.50 & 45.17 & 47 & 45.37 & 50.94
\\
\midrule 
\multicolumn{1}{l|}{BitDelta~\cite{liu2024bitdelta}} & 16$\times$ & 16$\times$ & 36.54 & 46.12 & 46.30 & 54.30 & 42.84 & 44.85 & 41.89 & 48.42
\\
\multicolumn{1}{l|}{Delta-CoMe~\cite{ping2024delta}} & 16$\times$ & 16$\times$ & 37.25 & 46.87 & 45.30 & 53.70 & 45.02 & 45.57 & 42.52 & 48.71
\\
\midrule 
\multicolumn{1}{l|}{DARE~\cite{yu2024language} (Sparsity)} & 14.7$\times$ & 23.7$\times$ & 36.62 & 46.70 & 46.20 & 53.90 & 33.54 & 36.29 & 38.79 & 45.63
\\
\multicolumn{1}{l|}{MP~\cite{han2015deep} (Sparsity)} & 14.7$\times$ & 23.7$\times$ & 23.65 & 3.11 & 45.70 & 47.60 & 40.02 & 38.25 & 36.46 & 29.65
\\
\multicolumn{1}{l|}{DARE~\cite{yu2024language} (Compression)} & 31.7$\times$ & 50.1$\times$ & 29.49 & 35.63 & 43.30 & 51.80 & 29.13 & 33.25 & 33.97 & 40.23
\\
\multicolumn{1}{l|}{MP~\cite{han2015deep} (Compression)} & 31.7$\times$ & 50.1$\times$ & 16.22 & 8.42 & 32.90 & 16.50 & 31.95 & 21.15 & 27.02 & 15.36
\\
\midrule 
\multicolumn{1}{l|}{\textbf{\methodname{} (Ours)}} & \textbf{32.9$\times$} & \textbf{50.9$\times$} & \textbf{38.76} & \textbf{48.45} & \textbf{51.80} & \textbf{59.10} & \textbf{46.14} & \textbf{48.59} & \textbf{45.57} & \textbf{52.05}
\\
\bottomrule
\end{tabular}
}

\end{table*}

%% file: tables/3_recent_llm.tex
\begin{table*}[htbp]
\centering
\caption{Performance comparison on recent large language models.}
\label{tab:performance_recent_llm} 
\resizebox{0.99\textwidth}{!}{%
\begin{tabular}{l|c|cc|cc|cc|c}
\toprule
\multirow{2}{*}{\textbf{Method}} 
& \multirow{2}{*}{\textbf{Ratio}} 
& \multicolumn{2}{c|}{\textbf{Llama-3.1-Tulu-3-8B}} 
& \multicolumn{2}{c|}{\textbf{Qwen2.5-7B-Instruct}} 
& \multicolumn{2}{c|}{\textbf{Qwen3Guard-Gen-8B}} 
& \multirow{2}{*}{\textbf{Avg}} \\
& & HumanEval+ & MBPP+ & MATH & GPQA & MMLU & BBQ & \\
\midrule
\multicolumn{1}{l|}{Individual} & 1$\times$ & 57.30 & 56.60 & 36.20 & 36.87 & 72.80 & 50.34 & 51.69 \\
\midrule 
\multicolumn{1}{l|}{BitDelta~\cite{liu2024bitdelta}} & 16$\times$ & 55.50 & 52.90 & 38.10 & 37.37 & 73.05 & 49.60 & 51.09 \\
\multicolumn{1}{l|}{Delta-CoMe~\cite{ping2024delta}} & 16$\times$ & 53.80 & 53.10 & 29.45 & 31.43 & 71.46 & 44.61 & 47.31 \\
\midrule 
\multicolumn{1}{l|}{DARE~\cite{yu2024language} (Sparsity)} & 14.7$\times$ & 50.60 & 50.00 & 37.76 & 34.85 & 72.18 & 49.69 & 49.18 \\
\multicolumn{1}{l|}{MP~\cite{han2015deep} (Sparsity)} & 14.7$\times$ & 45.10 & 40.20 & 11.84 & 31.82 & 70.91 & \textbf{50.18} & 41.68 \\
\midrule 
\multicolumn{1}{l|}{\textbf{\methodname{} (Ours)}} & \textbf{32.9$\times$} & \textbf{56.10} & \textbf{54.50} & \textbf{40.14} & \textbf{38.38} & \textbf{73.44} & 49.74 & \textbf{52.05} \\
\bottomrule
\end{tabular}
}
\end{table*}

%% file: tables/8_t5.tex
\begin{table*}[htbp]
\centering
\caption{Performance comparison on T5-base models across 8 tasks.}
\label{tab:performance_t5} 

\resizebox{0.99\textwidth}{!}{ 
\begin{tabular}{l|c|cccccccc|c}
\toprule
\textbf{Methods} & \textbf{Ratio} & \textbf{CoLA} & \textbf{SST2} & \textbf{MRPC} & \textbf{STSB} & \textbf{QQP} & \textbf{MNLI} & \textbf{QNLI} & \textbf{RTE} & \textbf{Avg} \\
\midrule
{Individual} & 1$\times$ & 74.98  &  93.58  &  87.50  &  88.70  & 85.37  &  83.41  &  91.49  &  85.92 &  86.37   \\
\midrule
{BitDelta~\cite{liu2024bitdelta}} & 16$\times$ &  70.09  &  93.46  &  84.06  & 86.20 &  85.26  &  \textbf{83.64}   &  90.94   & 83.75 & 84.68  \\
{Delta-CoMe~\cite{ping2024delta}} & 16$\times$ &  74.26  &  93.58  &  87.01  & 87.94 &  85.44  &  83.51   &  91.54   & 84.36 & 85.95  \\
\midrule
{DARE~\cite{yu2024language} (Sparsity)} & 127.6$\times$ & 74.16 & 93.34 & 87.83 & 87.86 & 85.22 & 83.19 & 91.52 & 84.11 & 85.90 \\
{MP~\cite{han2015deep} (Sparsity)} & 127.6$\times$ & 37.87 & 88.76 & 70.59 & 0 & 80.4 & 68.55 & 82.88 & 82.67 & 63.97 \\
{DARE~\cite{yu2024language} (Compression)} & 220.5$\times$ & 71.57 & 92.25 & 85.03 & 86.77 & 84.26 & 81.16 & 90.50 & 82.81 & 84.30 \\
{MP~\cite{han2015deep} (Compression)} & 220.5$\times$ & 30.87 & 87.84 & 70.83 & 0 & 79.16 & 56.36 & 76.31 & 82.17 & 60.44 \\
\midrule
\textbf{\methodname{} (Ours)} & \textbf{224.6$\times$} &  \textbf{76.51}  &  \textbf{93.81}  &  \textbf{88.48}  & \textbf{88.82} &  \textbf{85.41}  &  83.17   &  \textbf{91.76}   & \textbf{85.92} & \textbf{86.74}  \\

\bottomrule
\end{tabular}
}
\end{table*}

%% file: tables/8_vit_l_14.tex
\begin{table*}[htbp]
\centering
\caption{Performance comparison on ViT-L/14 models across 8 tasks.}
\label{tab:performance_vitlarge14} 

\resizebox{0.99\textwidth}{!}{ 
\begin{tabular}{l|c|cccccccc|c}
\toprule
\textbf{Methods} & \textbf{Ratio} & \textbf{SUN397} & \textbf{Cars} & \textbf{RESISC45} & \textbf{EuroSAT} & \textbf{SVHN} & \textbf{GTSRB} & \textbf{MNIST} & \textbf{DTD} & \textbf{Avg} \\
\midrule
{Individual}  & 1$\times$ & 84.8  &  92.3  &  97.4  &  99.7  &  98.1  &  99.2  &  99.7  &  84.1 & 94.4   \\
{Traditional MTL~\cite{caruana1997multitask}} & 8$\times$ & 80.8  &  90.6  &  96.3  & 96.3 &  97.6  & 99.1  &  99.6   & 84.4 & 93.1  \\
\midrule
{BitDelta~\cite{liu2024bitdelta}} & 16$\times$ &  84.0  &  92.1  &  97.2  & 99.7 &  97.9  &  99.0  &  99.7   & 83.0 & 94.1  \\
{Delta-CoMe~\cite{ping2024delta}} & 16$\times$ &  \textbf{84.3}  &  92.1  &  97.2  & 99.6 &  98.0  &  99.1   &  99.7   & 83.6 & 94.2  \\
\midrule
DARE~\cite{yu2024language} (Sparsity) & 66.5$\times$ & 82.2 & 90.0 & 97.0 & 99.7 & 98.0 & 99.1 & 99.7 & 83.1 & 93.6 \\
MP~\cite{han2015deep} (Sparsity)   & 66.5$\times$ & 26.7 & 43.9 & 83.4 & 92.8 & 24.4 & 62.9 & 77.7 & 66.1 & 59.7 \\
DARE~\cite{yu2024language} (Compression) & 127.6$\times$ &  65.2 & 78.4 & 94.2 & 99.6 & 97.7 & 99.2 & 99.7 & 79.1 & 89.1 \\
MP~\cite{han2015deep} (Compression)   & 127.6$\times$ &  4.9 &  4.2 &  42.6 & 83.2 &  6.4 &  16.7 &  27.2 & 50.7 & 29.5 \\
\midrule
\textbf{\methodname{} (Ours)} & \textbf{132.5$\times$} &  84.1  &  \textbf{92.2}  &  \textbf{97.4}  & \textbf{99.8} &  \textbf{98.1}  &  \textbf{99.2}   &  \textbf{99.8}   & \textbf{84.3} & \textbf{94.4}  \\
\bottomrule

\end{tabular}
}
\end{table*}

%% file: tables/3_beit3.tex
\begin{table*}[htbp]

\centering
\caption{Performance comparison on multi-modal BEiT-3 models across 3 vision-language tasks.}
\label{tab:performance_beit} 
\resizebox{0.99\textwidth}{!}{
\begin{tabular}{lc|c|cccc|c|c}
\toprule

\multirow{2}{*}{\textbf{Methods}} &\multicolumn{1}{|c|}{\textbf{Task}} & \textbf{Ratio} & \multicolumn{4}{c|}{\textbf{COCO-Captioning}} & \textbf{NLVR2}& \textbf{VQAv2}
\\
& \multicolumn{1}{|c|}{Metric} & $\alpha(\uparrow)$ & BLEU4($\uparrow$) & CIDEr($\uparrow$) & METEOR($\uparrow$) & ROUGE-L($\uparrow$) & Accuracy($\uparrow$) & Accuracy($\uparrow$)

\\
\midrule

\multicolumn{2}{l|}{Individual} & 1$\times$ & 0.394 & 1.337 & 0.331 & 0.601 & 84.269 & 83.49
\\
\midrule 
\multicolumn{2}{l|}{BitDelta~\cite{liu2024bitdelta}} & 16$\times$ & 0.367 & 1.250 & 0.294 & 0.581 & 82.317 & 74.59 
\\
\multicolumn{2}{l|}{Delta-CoMe~\cite{ping2024delta}} & 16$\times$ & 0.224 & 0.846 & 0.251 & 0.488 & 60.098 & 38.64 
\\
\midrule 
\multicolumn{2}{l|}{DARE~\cite{yu2024language} (Sparsity)} & 7.7$\times$ & 0.333 & 1.15 & 0.285 & 0.564 & 80.652 & 73.29 \\
\multicolumn{2}{l|}{MP~\cite{han2015deep} (Sparsity)} & 7.7$\times$ & 0 & 0 & 0.006 & 0.014 & 49.263 & 0 \\
\multicolumn{2}{l|}{DARE~\cite{yu2024language} (Compression)} & 18.1$\times$ & 0.178 & 0.589 & 0.196 & 0.441 & 69.470 & 58.89 \\
\multicolumn{2}{l|}{MP~\cite{han2015deep} (Compression)} & 18.1$\times$ & 0 & 0 & 0.006 & 0.013 & 48.931 & 0.01 \\
\midrule 
\multicolumn{2}{l|}{\textbf{\methodname{} (Ours)}} & \textbf{18.4$\times$} & \textbf{0.372} & \textbf{1.270} & \textbf{0.296} & \textbf{0.583} & \textbf{83.163} & \textbf{75.66}
\\
\bottomrule
\end{tabular}
}

\end{table*}

%% file: tables/abl_component.tex
\begin{table*}[t]
\centering
\caption{Ablation study on the effectiveness of each component.}
\label{tab:ablation_component}
\resizebox{0.99\textwidth}{!}{
\begin{tabular}{l|cccccccc|c|c}
\toprule
\textbf{Methods} & \textbf{SUN397} & \textbf{Cars} & \textbf{RESISC45} & \textbf{EuroSAT} & \textbf{SVHN} & \textbf{GTSRB} & \textbf{MNIST} & \textbf{DTD} & \textbf{Ratio} & \textbf{Avg} \\
\midrule
DARE~\cite{yu2024language} (97\% Sparsity) & 76.0 & 73.3 & 95.1 & 99.7 & 97.0 & 98.5 & 99.6 & 78.1 & 23.7$\times$ & 89.7 \\
\midrule
+ \pruningshort{} & 75.6 & 73.3 & 95.3 & 99.5 & 97.1 & 98.5 & 99.6 & 78.5 & 50.9$\times$ ($\uparrow$ 27.2$\times$) & 89.7 \\
+ \pruningshort{} + \allocationshort{} & 76.9 & 75.9 & 95.5 & 99.2 & 97.0 & 98.8 & 99.7 & 79.2 & 50.9$\times$ & 90.3 ($\uparrow$ 0.6) \\
+ \pruningshort{} + \allocationshort{} + \rescalingshort{} & 78.6 & 77.4 & 95.7 & 99.4 & 97.0 & 98.7 & 99.7 & 79.0 & 50.9$\times$ & 90.7 ($\uparrow$ 0.4) \\
\bottomrule
\end{tabular}
}
\end{table*}

%% file: tables/abl_storage.tex
\begin{table*}[t]
\centering
\caption{Compression ratios across storage formats and models.}
\label{tab:storage_formats_transposed}
\resizebox{0.99\textwidth}{!}{
\begin{tabular}{l|ccccccc}
\toprule
\textbf{Format} & \textbf{LLaMA-2-7B} & \textbf{LLaMA-2-13B} & \textbf{ViT-B/32} & \textbf{ViT-L/14} & \textbf{T5-base} & \textbf{RoBERTa-base} & \textbf{BEiT-3} \\
\midrule
Index-Free Storage & 80.0$\times$  & 133.0$\times$ & 133.0$\times$ & 400.0$\times$ & 800.0$\times$ & 80.0$\times$   & 40.0$\times$ \\
Golomb Coding & 32.9$\times$  & 50.9$\times$  & 50.9$\times$  & 132.5$\times$ & 224.6$\times$ & 32.9$\times$   & 18.4$\times$ \\
CSR  & 19.5$\times$ & 30.7$\times$ & 35.4$\times$ & 105.2$\times$& 188.3$\times$& 21.4$\times$  & 10.8$\times$ \\
BCSR & 12.7$\times$ & 19.8$\times$ & 21.7$\times$ & 63.7$\times$& 121.4$\times$& 13.4$\times$  & 7.3$\times$ \\
\bottomrule
\end{tabular}
}
\end{table*}

%% file: paragraphs/5_discussion.tex
\section{Discussion and Conclusion}
\textbf{Regularization Effect.} 
By controlling the fitting degree of models, we examine that \methodname{} particularly benefits underfitted models (see App.~\ref{appendix:discussioin_regularization} for detailed results and analysis). 
Since LLMs are typically evaluated in the zero-shot or few-shot settings, where underfitting is common, this explains why \methodname{} often yields not only strong compression but also performance gains.

\textbf{Overhead.} 
\methodname{} introduces only modest computational overhead during compression and is efficient at inference compared with baselines. 
The main cost arises from computing trace norms. 
Since only their relative magnitudes are required, fast approximation such as randomized SVD can be employed to reduce the overhead.
In the inference stage, delta weights only need to be dequantized and added back to the base model, which can be done on CPU efficiently.
In contrast, methods such as Delta-CoMe~\cite{ping2024delta} require additional matrix multiplications, making them less efficient.

\textbf{Limitations and Future Work.} We acknowledge that our method employs a heuristic rescaling strategy and does not explicitly target deployment efficiency; detailed discussion is in App.~\ref{appendix:discussion_limitation}.

\textbf{Conclusion.}
In this paper, we present \methodname{}, the first data-free pipeline for ultra-efficient delta compression, capable of achieving both ultra-high compression ratios and strong performance.
\methodname{} is a hybrid compression method with a focus on preserving information across inter-layer, intra-layer, and global dimensions through three novel components. Extensive experiments on language, vision, and multi-modal models, along with theoretical analysis, validate the effectiveness, robustness, and generality of \methodname{} across diverse settings.

%% file: paragraphs/appendix.tex
\begin{center}
    \LARGE \textbf{Appendix for \methodname{}}
\end{center}
\vspace{1em}

\section{Non-quantized Variant of \pruningname{}}
\label{appendix:noquant_das}
While our main \pruningname{} operates on quantized delta weights, it can also be adapted to a non-quantized variant for scenarios where quantization is not applied or not desired. In this version, we replace the discrete grouping based on quantized values with continuous interval-based grouping over the full-precision delta weights.

Specifically, for a given delta weight $\Delta\theta_t$, we first compute its minimum and maximum values, denoted as $\min(\Delta\theta_t)$ and $\max(\Delta\theta_t)$. We then divide the range $[\min(\Delta\theta_t), \max(\Delta\theta_t)]$ into $I$ equally spaced intervals. The width of each interval is given by:

\begin{equation}
\delta_{\text{interval}} = \frac{\max(\Delta\theta_t) - \min(\Delta\theta_t)}{I}
\end{equation}

The $i$-th interval is then defined as:
\begin{equation}
\mathcal{I}i = \left[\min(\Delta\theta_t) + (i-1)\cdot \delta_{\text{interval}}, \min(\Delta\theta_t) + i \cdot \delta_{\text{interval}}\right)], \quad \text{for } i = 1, \ldots, I
\end{equation}

To perform group-wise pruning, we define a binary mask
for each interval $I_i$, where pruning decisions are sampled from a Bernoulli distribution with success probability $1-s_l$, and $s_l$ is the sparsity rate of the layer. Specifically:
\begin{equation}
\boldsymbol{M}_{I_i}^{(j,k)} \sim 
\begin{cases}
\mathrm{Bernoulli}(1 - s_l), & \text{if } {\Delta\theta_t}^{(j,k)} \in I_i \\
0, & \text{otherwise}
\end{cases}
\label{eq:mask_definition2}
\end{equation}
The pruned delta weight ${\Delta\theta_t}^*$ is obtained by applying the group-wise Bernoulli masks to the quantized delta weights:
\begin{equation}
\Delta\theta_t^* = \sum_{i=1}^I \boldsymbol{M}_{\mathcal{I}_i} \odot \Delta\theta_t
\end{equation}

\section{Detailed Derivation}
\subsection{Derivation of the Variance-Entropy Relationship}
\label{appendix:variance_entropy}
We aim to derive the relationship between the variance and the entropy of a layer.
First, we define the entropy of a random variable $L$ with probability density function $p(l)$ as:
\begin{equation}
H(L) = -\mathbb{E}[\log p(l)]
\end{equation}

Assuming $L$ follows a Gaussian distribution $\mathcal{N}(\mu, \sigma^2)$, its probability density function is:
\begin{equation}
p(l) = \frac{1}{\sqrt{2\pi\sigma^2}} \exp\left(-\frac{(l - \mu)^2}{2\sigma^2} \right)
\end{equation}

Taking the logarithm of the density function, we obtain:
\begin{equation}
\log p(l) = \log\left( \frac{1}{\sqrt{2\pi\sigma^2}} \exp\left(-\frac{(l - \mu)^2}{2\sigma^2} \right) \right) 
= -\frac{1}{2} \log(2\pi\sigma^2) - \frac{(l - \mu)^2}{2\sigma^2}
\end{equation}

Substituting this into the definition of entropy, we have:
\begin{align}
H(L) 
&= -\int_{-\infty}^{\infty} p(l) \left( -\frac{1}{2} \log(2\pi\sigma^2) - \frac{(l - \mu)^2}{2\sigma^2} \right) dl \notag \\
&= \frac{1}{2} \log(2\pi\sigma^2) \int p(l) dl + \frac{1}{2\sigma^2} \int p(l)(l - \mu)^2 dl
\end{align}

Next, we use known properties of the Gaussian distribution:
\begin{equation}
\int p(l) dl = 1 
\quad \text{and} \quad 
\int p(l)(l - \mu)^2 dl = \mathrm{Var}(L) = \sigma^2
\end{equation}

Substituting these into the expression for entropy yields:
\begin{align}
H(L) 
&= \frac{1}{2} \log(2\pi\sigma^2) + \frac{1}{2} \notag \\
&= \frac{1}{2} \log(2\pi) + \log(\sigma) + \frac{1}{2}
\end{align}

Therefore, we can conclude that the entropy $H(L)$ increases logarithmically with the standard deviation $\sigma$ of the distribution. Since the variance is $\sigma^2$, entropy is monotonically increasing with variance. This derivation shows that the entropy of a layer is positively correlated with its variance. Since entropy represents the amount of information, layers with larger variance contain more information. As a result, such layers should be assigned a smaller sparsity rate to preserve more information.

\subsection{Derivation of the Variance of Activation Error}
\label{appendix:activation_error}
According to Eq.~\ref{equation:activation_error}, the activation error $\epsilon$ is defined as:
\begin{equation}\varepsilon=a\odot\left(1-\frac{\gamma}{1-s}B\right),\quad\mathrm{where~}a=\Delta\theta\odot x\end{equation}

Here, $B$ is a Bernoulli random variable, where the probability of success is $1-s$. For a Bernoulli distribution, we know that the expectation and second moment are $\mathbb{E}[B]=1-s$ and $\mathbb{E}[B^2]=1-s$.
We first calculate the expectation of $\epsilon$. Using the linearity of expectation:
\begin{equation}\begin{aligned}
\mathbb{E}[\varepsilon] & =a\odot\mathbb{E}\left[1-\frac{\gamma}{1-s}B\right] \\
 & =a\odot\left(1-\frac{\gamma}{1-s}\mathbb{E}[B]\right) \\
 & =a\odot\left(1-\frac{\gamma}{1-s}(1-s)\right) \\
 & =a\odot(1-\gamma)
\end{aligned}\end{equation}

Next, we calculate the expectation of $\epsilon^2$:
\begin{equation}\begin{aligned}
\mathbb{E}[\varepsilon^{2}] & =a^2\odot\mathbb{E}\left[1-2\frac{\gamma}{1-s}B+\left(\frac{\gamma}{1-s}B\right)^2\right] \\
 & =a^2\odot\left[1-2\frac{\gamma}{1-s}\mathbb{E}[B]+\left(\frac{\gamma}{1-s}\right)^2\mathbb{E}[B]\right] \\
 & =a^2\odot\left[1-2\frac{\gamma}{1-s}(1-s)+\left(\frac{\gamma}{1-s}\right)^2(1-s)\right] \\
 & =a^2\odot\left(1-2\gamma+\frac{\gamma^2}{1-s}\right)
\end{aligned}\end{equation}

Now, we can calculate the variance of $\epsilon$. The variance is given by:
\begin{equation}\begin{aligned}
\mathrm{Var}(\varepsilon) & =\mathbb{E}[\varepsilon^2]-(\mathbb{E}[\varepsilon])^2 \\
 & =a^2\odot\left(1-2\gamma+\frac{\gamma^2}{1-s}\right)-a^2\odot(1-\gamma)^2 \\
 & =\boldsymbol{a}^2\odot\left[\left(1-2\gamma+\frac{\gamma^2}{1-s}\right)-(1-2\gamma+\gamma^2)\right] \\
 & =a^2\odot\left(\frac{\gamma^2s}{1-s}\right)
\end{aligned}\end{equation}

From the derived variance expression, we can observe that when the sparsity $s$ is high, the variance will become very large, leading to unstable performance. To mitigate this, we can choose a value for $\gamma$ in the range $[0,1]$ to reduce the impact of the original rescaling factor. However, $\gamma$ cannot be chosen too small. This is because the expectation of the activation error is $a\odot(1-\gamma)$, and if $\gamma$ is too small, the expectation of the activation error increases, which may lead to undesirable effects. Therefore, $\gamma$ is typically set in the range $[0.5,1.0]$ under high sparsity, which balances both the variance and the expectation of the activation error.

\subsection{Detailed Derivation of Compression Ratio}
\label{appendix:golomb}
\textbf{Golomb Coding.} 
Golomb coding~\cite{golomb1966run} is a run-length entropy coding scheme particularly well-suited for a geometric distribution, where it encodes the distances between successive non-zero entries (i.e., run-lengths of zeros). Unlike conventional formats such as Compressed Sparse Row (CSR), which store explicit indices or pointers for non-zero values, run-length coding represents zero runs instead of storing indices, which can yield a more compact representation than CSR when sparsity is high.

\textbf{Compression Ratio Derivation.} 
In our setting, after pruning, most entries in the delta weight are zeros, and the positions of non-zeros can be modeled as a Bernoulli process with success probability equal to the density $1-s$, where $s$ is the sparsity. Consequently, the run-lengths of zeros follow a geometric distribution, for which Golomb coding achieves an average code length close to the entropy of the underlying geometric distribution. The entropy of a geometric distribution is:
\begin{equation}
    H_{\text{geo}} = - (1-s)\log_2(1-s) - s\log_2 s
\end{equation}

When each non-zero entry can take $m$ discrete values, the maximum entropy contribution of its symbol is $\log_2 m$, which corresponds to the case where all $m$ values are equally likely.
In practice, the values of our non-zero entries follow a truncated Gaussian distribution, so the actual entropy is smaller than $\log_2 m$, implying that the achievable compression ratio is higher. 

For simplicity of analysis, we approximate this term by $\log_2 m$, which serves as an upper bound, and the per-parameter entropy of the compressed model is:

\begin{equation}
    H_{\text{comp}} \approx -\Big[s\log_2 s + (1-s)\log_2 \tfrac{(1-s)}{m}\Big]
\end{equation}

By contrast, the original uncompressed model stored in FP16 format requires 
$H_{\text{orig}} = 16$ bits per parameter. Therefore, the theoretical compression ratio is:
\begin{equation}
\label{equation:cr}
    \text{CR} = \frac{H_{\text{orig}}}{H_{\text{comp}}} 
\approx \frac{16}{- \big[s\log_2 s + (1-s)\log_2 \tfrac{(1-s)}{m}\big]}
\end{equation}

Note that the rescaling factor is represented by a single 16-bit value, which is negligible compared to the overall storage and is therefore ignored when reporting the compression ratio. 

\textbf{Detailed Compression Ratios.}
We also report the compression ratios and corresponding sparsity rates of DARE~\cite{yu2024language}, MP~\cite{han2015deep}, and \methodname{} under each setting. Practical compression ratios with Golomb coding are computed following Eq.~\ref{equation:cr}, using $m=16$ for DARE and MP (16-bit precision) and $m=4$ for \methodname{} (4-bit quantization). For completeness, we also include Index-Free compression ratios, which denote the ideal upper bound where only non-zero values are stored. The results are shown in Tab.~\ref{tab:perf_all}.
\input{tables/ideal_practical}

\section{Additional Experimental Settings and Results}
\subsection{HuggingFace IDs for LLM Checkpoints}
\label{appendix:model_ids}
We provide the Hugging Face IDs of the LLMs used in our experiments, as shown in Tab.~\ref{tab:llm_ids}. All models are evaluated using the lm-evaluation-harness~\cite{eval-harness} framework for general language tasks and EvalPlus~\cite{evalplus} framework for code generation tasks.
\input{tables/llm_ids}

\subsection{Experiments on eight ViT-B/32 models}
\label{appendix:vitb_results}

\textbf{Results.} The results of ViT-B/32 models are shown in Tab.~\ref{tab:performance_vitbase32}.
With 4-bit quantization and 97\% sparsity, \methodname{} achieves 50.9$\times$ compression ratio on ViT-B/32.
\methodname{} achieves an average accuracy of 90.7 across 8 diverse vision tasks, nearly matching the performance of individually fine-tuned models (91.0). Compared to all baselines, \methodname{} achieves both higher compression and better accuracy.

\input{tables/8_vit_b_32}

\subsection{Experiments on eight RoBERTa-base models}
\label{appendix:roberta_results}

\textbf{Settings.} For RoBERTa-base, we report Matthews correlation on CoLA, the average of Pearson and Spearman correlations on STS-B, and accuracy on the other tasks.

\textbf{Results.} The results of RoBERTa-base models are shown in Tab.~\ref{tab:performance_roberta}.
For RoBERTa, we apply 4-bit quantization and prune 95\% of parameters, achieving an overall 32.9$\times$ compression ratio.
Despite the high compression, \methodname{} achieves an average score of 84.46, closely matching the uncompressed individually fine-tuned model (85.56) and outperforming all baselines by a notable margin.

\input{tables/8_roberta}

\subsection{Detailed Results of Ablation Study}
\label{appendix:ablation}

\subsubsection{Ablation on \pruningname{}}
\label{appendix:ablation_dac}

We present the detailed numerical results of the ablation study conducted on \pruningshort{}. For brevity, we report only the results under 90\% sparsity, as summarized in Tab.~\ref{tab:30_vit}.
Following~\cite{huang2024emr,zheng2024free}, we use ViT-B/16 as the pre-trained backbone model and evaluate its performance across 30 diverse image classification datasets. Specifically, the datasets are: MNIST~\cite{lecun1998mnist},
CIFAR-10~\cite{krizhevsky2009learning},
CIFAR-100~\cite{krizhevsky2009learning},
Cars~\cite{krause20133d},
Fashion-MNIST~\cite{xiao2017fashion},
EMNIST~\cite{cohen2017emnist},
STL10~\cite{coates2011analysis},
GTSRB~\cite{stallkamp2011german},
SHVN~\cite{yuval2011reading},
Oxford-IIIT Pet~\cite{parkhi2012cats},
Cats and Dogs~\cite{cukierskidogs},
Dogs~\cite{KhoslaYaoJayadevaprakashFeiFei_FGVC2011},
Beans~\cite{beansdata},
Food-101~\cite{bossard14},
Fruits-360~\cite{muresan2018fruit},
Vegetables~\cite{ahmed2021dcnn}, 
MangoLeafBD~\cite{ahmed2023mangoleafbd},
Flowers Recognition~\cite{Flowers},
Landscape Recognition~\cite{Landscape},
Weather~\cite{xiao2021classification},
DTD~\cite{cimpoi2014describing},
EuroSAT~\cite{helber2019eurosat}, 
RESISC45~\cite{cheng2017remote},
SUN397~\cite{xiao2010sun},
KenyanFood13~\cite{jalal2019scraping}, 
Intel Images~\cite{bansal2019intel},
Garbage Classification~\cite{cchang_2018},
Animal-10N~\cite{song2019selfie},
CUB-200-2011~\cite{wah2011caltech},
and Kvasir-v2~\cite{pogorelov2017kvasir}.
These datasets are chosen to comprehensively evaluate generalization across various data distributions and classification challenges. Accuracy is used as the evaluation metric for all datasets.

\input{tables/30_vit_b_16}

\subsubsection{Ablation on Hyper-Parameters}
\label{appendix:ablation_hyper}
We present the detailed numerical results of our hyper-parameter ablation studies in Tab.~\ref{tab:ablation_hyperparameter}. The results complement the analysis provided in the main text and offer full visibility into the performance under different settings.
\input{tables/abl_hyperparameter}

\section{Discussion}
\label{appendix:discussion}

\subsection{Regularization Effect}
\label{appendix:discussioin_regularization}
We observe in experiments that \methodname{} sometimes even outperforms fine-tuned models, especially in the LLM setting. This can be attributed to overfitting or underfitting during fine-tuning.  

To verify this, we conducted controlled experiments on two datasets, SUN397~\cite{xiao2010sun} and Cars~\cite{krause20133d}, using a pretrained ViT-B/32 model. By varying the number of fine-tuning steps, we controlled the degree of model fitting. The results are shown in Tab.~\ref{tab:control_exp}, and we find that underfitted models (early training steps, higher loss) consistently benefit from \methodname{}, while well-trained models (later steps, lower loss) show some or no improvement. These results indicate that \methodname{} mainly benefits underfitted models, while well-trained models already capture most task-specific information and thus leave little room for further improvement.

This also explains why improvements are more common in LLMs, which are typically evaluated in zero-shot or few-shot settings.
In such scenarios, these models are not explicitly fine-tuned on the target tasks, which often leads to weaker generalization. 
As a result, they can be regarded as underfitted to the evaluation benchmarks. 
In this context, \methodname{} provides a regularization effect, enabling underfitted models to benefit more from compression.

\input{tables/controlled_experiment}

\subsection{Compression Ratio Differences Across Models}
We observe that different models can be compressed to different extents without losing performance. This variation mainly comes from the amount of redundancy in their delta weights. For instance, T5-base has relatively high redundancy, making it possible to compress it up to 224$\times$ while maintaining or even improving performance. In contrast, models like BEiT-3 have delta weights with larger magnitudes and lower redundancy, leaving less room for aggressive compression. This suggests that the achievable compression ratio depends not only on the compression method itself, but also on the model architecture, its training process, and the nature of the fine-tuned tasks.

\subsection{Limitations and Future Work}
\label{appendix:discussion_limitation}
\textbf{Heuristic Rescaling Factor.} 
One limitation of our method lies in the heuristic nature of the additional rescaling factor $\gamma$. 
We set $\gamma$ according to the trace norm of each delta weight, which serves as a lightweight proxy for activation instability under the data-free setting. 
While this simple strategy helps mitigate instability at high sparsity, it remains a heuristic and may not be optimal.
Exploring more principled or adaptive approaches for determining $\gamma$ under data-free constraints remains an important direction for future work.

\textbf{Deployment Efficiency.} 
Another limitation is that our method is not designed to directly optimize deployment efficiency, such as inference acceleration. 
Achieving real speedup typically requires specialized GPU kernels or structured sparsity to better exploit hardware parallelism. 
Future work may combine our approach with optimized kernels or structured pruning to realize both strong compression–performance trade-offs and practical deployment gains.

%% file: tables/ideal_practical.tex
\begin{table*}[t]
\centering
\caption{Compression ratios, average performance, and corresponding sparsity rates across different models. ``Practical'' denotes the practical compression ratio using Golomb coding, ``Ideal'' the ideal upper bound, and ``Avg'' the average performance.}
\label{tab:perf_all}
\resizebox{\textwidth}{!}{
\begin{tabular}{l|cccc|cccc}
\toprule
\multirow{2}{*}{\textbf{Method}} 
& \multicolumn{4}{c|}{\textbf{LLaMA-2-7B (Tab.~\ref{tab:performance_llm})}} 
& \multicolumn{4}{c}{\textbf{LLaMA-2-13B (Tab.~\ref{tab:performance_llm})}} \\
& Practical & Ideal & Avg & Sparsity 
& Practical & Ideal & Avg & Sparsity \\
\midrule
DARE~\cite{yu2024language} (Sparsity) 
& 14.7$\times$ & 20$\times$ & 38.79 & 95.0\% 
& 23.7$\times$ & 33.3$\times$ & 45.63 & 97.0\% \\
MP~\cite{han2015deep} (Sparsity)   
& 14.7$\times$ & 20$\times$ & 36.46 & 95.0\% 
& 23.7$\times$ & 33.3$\times$ & 29.65 & 97.0\% \\
\midrule
DARE~\cite{yu2024language} (Compression) 
& 31.7$\times$ & 45.5$\times$ & 33.97 & 97.8\% 
& 50.1$\times$ & 74.1$\times$ & 40.23 & 98.65\% \\
MP~\cite{han2015deep} (Compression)   
& 31.7$\times$ & 45.5$\times$ & 27.02 & 97.8\% 
& 50.1$\times$ & 74.1$\times$ & 15.36 & 98.65\% \\
\midrule
\methodname{} (Ours) 
& \textbf{32.9$\times$} & \textbf{80.0$\times$} & \textbf{45.57} & 95.0\% 
& \textbf{50.9$\times$} & \textbf{133.0$\times$} & \textbf{52.05} & 97.0\% \\
\midrule\midrule

\multirow{2}{*}{\textbf{Method}} 
& \multicolumn{4}{c|}{\textbf{Recent LLMs (Tab.~\ref{tab:performance_recent_llm})}} 
& \multicolumn{4}{c}{\textbf{ViT-B/32 (Tab.~\ref{tab:performance_vitbase32})}} \\
& Practical & Ideal & Avg & Sparsity 
& Practical & Ideal & Avg & Sparsity \\
\midrule
DARE~\cite{yu2024language} (Sparsity) 
& 14.7$\times$ & 20$\times$ & 49.18 & 95.0\% 
& 23.7$\times$ & 33.3$\times$ & 89.7 & 97.0\% \\
MP~\cite{han2015deep} (Sparsity)   
& 14.7$\times$ & 20$\times$ & 41.68 & 95.0\% 
& 23.7$\times$ & 33.3$\times$ & 77.4 & 97.0\% \\
\midrule
DARE~\cite{yu2024language} (Compression) 
& -- & -- & -- & -- 
& 50.1$\times$ & 74.1$\times$ & 83.5 & 98.65\% \\
MP~\cite{han2015deep} (Compression)   
& -- & -- & -- & -- 
& 50.1$\times$ & 74.1$\times$ & 53.1 & 98.65\% \\
\midrule
\methodname{} (Ours) 
& \textbf{32.9$\times$} & \textbf{80.0$\times$} & \textbf{52.05} & 95.0\% 
& \textbf{50.9$\times$} & \textbf{133.0$\times$} & \textbf{90.7} & 97.0\% \\
\midrule\midrule

\multirow{2}{*}{\textbf{Method}} 
& \multicolumn{4}{c|}{\textbf{ViT-L/14 (Tab.~\ref{tab:performance_vitlarge14})}}
& \multicolumn{4}{c}{\textbf{T5-base (Tab.~\ref{tab:performance_t5})}} \\
& Practical & Ideal & Avg & Sparsity 
& Practical & Ideal & Avg & Sparsity \\
\midrule
DARE~\cite{yu2024language} (Sparsity) 
& 66.5$\times$ & 100.0$\times$ & 93.6 & 99.0\% 
& 127.6$\times$ & 200.0$\times$ & 85.90 & 99.5\% \\
MP~\cite{han2015deep} (Sparsity)   
& 66.5$\times$ & 100.0$\times$ & 59.7 & 99.0\% 
& 127.6$\times$ & 200.0$\times$ & 63.97 & 99.5\% \\
\midrule
DARE~\cite{yu2024language} (Compression) 
& 127.6$\times$ & 200.0$\times$ & 89.1 & 99.5\% 
& 220.5$\times$ & 357.0$\times$ & 84.30 & 99.72\% \\
MP~\cite{han2015deep} (Compression)   
& 127.6$\times$ & 200.0$\times$ & 29.5 & 99.5\% 
& 220.5$\times$ & 357.0$\times$ & 60.44 & 99.72\% \\
\midrule
\methodname{} (Ours) 
& \textbf{132.5$\times$} & \textbf{400.0$\times$} & \textbf{94.4} & 99.0\% 
& \textbf{224.6$\times$} & \textbf{800.0$\times$} & \textbf{86.74} & 99.5\% \\
\midrule\midrule

\multirow{2}{*}{\textbf{Method}} 
& \multicolumn{4}{c|}{\textbf{RoBERTa-base (Tab.~\ref{tab:performance_roberta})}} 
& \multicolumn{4}{c}{\textbf{BEiT-3 (Tab.~\ref{tab:performance_beit})}} \\
& Practical & Ideal & Avg & Sparsity 
& Practical & Ideal & Avg & Sparsity \\
\midrule
DARE~\cite{yu2024language} (Sparsity) 
& 14.7$\times$ & 20.0$\times$ & 83.23 & 95.0\% 
& 7.7$\times$ & 10.0$\times$ & 70.74 & 90.0\% \\
MP~\cite{han2015deep} (Sparsity)   
& 14.7$\times$ & 20.0$\times$ & 80.06 & 95.0\% 
& 7.7$\times$ & 10.0$\times$ & 16.59 & 90.0\% \\
\midrule
DARE~\cite{yu2024language} (Compression) 
& 31.7$\times$ & 45.5$\times$ & 73.12 & 97.8\% 
& 18.1$\times$ & 25.0$\times$ & 54.49 & 96.0\% \\
MP~\cite{han2015deep} (Compression)   
& 31.7$\times$ & 45.5$\times$ & 69.87 & 97.8\% 
& 18.1$\times$ & 25.0$\times$ & 16.47 & 96.0\% \\
\midrule
\methodname{} (Ours) 
& \textbf{32.9$\times$} & \textbf{80.0$\times$} & \textbf{84.46} & 95.0\% 
& \textbf{18.4$\times$} & \textbf{40.0$\times$} & \textbf{73.95} & 90.0\% \\
\bottomrule
\end{tabular}
}
\end{table*}

%% file: tables/llm_ids.tex
\begin{table}[ht]
\centering
\caption{LLM checkpoint IDs.}
\label{tab:llm_ids}
\begin{small}
\resizebox{0.97\linewidth}{!}{%
\begin{tabular}{l|cc}
\toprule
\textbf{Models} & Pre-trained & Fine-tuned \\
\midrule
\multicolumn{3}{l}{\textbf{Model Size 7B (Llama-2 Series)}} \\
\midrule
Chat & \textit{meta-llama/Llama-2-7b-hf} & \textit{meta-llama/Llama-2-7b-chat-hf} \\
Code & \textit{codellama/CodeLlama-7b-Python-hf} & \textit{vanillaOVO/WizardCoder-Python-7B-V1.0} \\
Math & \textit{meta-llama/Llama-2-7b-hf} & \textit{WizardLMTeam/WizardMath-7B-V1.0} \\
\midrule
\multicolumn{3}{l}{\textbf{Model Size 13B (Llama-2 Series)}} \\
\midrule
Chat & \textit{meta-llama/Llama-2-13b-hf} & \textit{meta-llama/Llama-2-13b-chat-hf} \\
Code & \textit{codellama/CodeLlama-13b-Python-hf} & \textit{WizardLMTeam/WizardCoder-Python-13B-V1.0} \\
Math & \textit{meta-llama/Llama-2-13b-hf} & \textit{vanillaOVO/WizardMath-13B-V1.0} \\
\midrule
\multicolumn{3}{l}{\textbf{Recent Models}} \\
\midrule
Llama-3.1 & \textit{meta-llama/Llama-3.1-8B} & \textit{allenai/Llama-3.1-Tulu-3-8B-SFT} \\
Qwen2.5   & \textit{Qwen/Qwen2.5-7B} & \textit{Qwen/Qwen2.5-7B-Instruct} \\
Qwen3     & \textit{Qwen/Qwen3-8B} & \textit{Qwen/Qwen3Guard-Gen-8B} \\
\bottomrule
\end{tabular}
} 
\end{small}
\end{table}

%% file: tables/8_vit_b_32.tex
\begin{table*}[htbp]
\centering
\caption{Performance comparison on ViT-B/32 models across 8 tasks.}
\label{tab:performance_vitbase32} 

\resizebox{0.99\textwidth}{!}{ 
\begin{tabular}{l|c|cccccccc|c}
\toprule
\textbf{Methods} & \textbf{Ratio} & \textbf{SUN397} & \textbf{Cars} & \textbf{RESISC45} & \textbf{EuroSAT} & \textbf{SVHN} & \textbf{GTSRB} & \textbf{MNIST} & \textbf{DTD} & \textbf{Avg} \\
\midrule
{Individual} & 1$\times$ & 79.2  &  77.7  &  96.1  &  99.8  &  97.4  &  98.7  &  99.7  &  79.4 & 91.0   \\
{Traditional MTL~\cite{caruana1997multitask}} & 8$\times$ & 73.9  &  74.4  &  93.9  & 98.2 &  95.8  &  98.9   &  99.5   & 77.9 & 89.1  \\
\midrule
{BitDelta~\cite{liu2024bitdelta}} & 16$\times$ &  78.3  &  75.9  &  95.4  & 99.3 &  96.4  &  98.2   &  99.2   & 78.4 & 90.1  \\
{Delta-CoMe~\cite{ping2024delta}} & 16$\times$ &  78.5  &  72.3  &  95.6  & 99.6 &  \textbf{97.3}  &  98.6   &  99.5   & 78.3 & 90.0  \\
\midrule
DARE~\cite{yu2024language} (Sparsity) & 23.7$\times$ & 76.0 & 73.3 & 95.1 & \textbf{99.7} & 97.0 & 98.5 & 99.6 & 78.1 & 89.7 \\
MP~\cite{han2015deep} (Sparsity) & 23.7$\times$ & 58.1 & 44.2 & 91.2 & 96.1 & 91.3 & 73.5 & 98.3 & 66.7 & 77.4 \\
DARE~\cite{yu2024language} (Compression) & 50.1$\times$ & 57.2 & 55.2 & 91.3 & 99.4 & 96.4 & 98.2 & 99.6 & 71.1 & 83.5 \\
MP~\cite{han2015deep} (Compression) & 50.1$\times$ & 18.8 & 16.4 & 81.8 & 88.3 & 70.3 & 28.5 & 77.8 & 42.7 & 53.1 \\
\midrule
\textbf{\methodname{} (Ours)} & \textbf{50.9$\times$} &  \textbf{78.6}  &  \textbf{77.4}  &  \textbf{95.7}  & 99.4 &  97.0  &  \textbf{98.7}   &  \textbf{99.7}   & \textbf{79.0} & \textbf{90.7}  \\
\bottomrule
\end{tabular}
}
\end{table*}

%% file: tables/8_roberta.tex
\begin{table*}[htbp]
\centering
\caption{Performance comparison on RoBERTa-base models across 8 tasks.}
\label{tab:performance_roberta} 

\resizebox{0.99\textwidth}{!}{ 
\begin{tabular}{l|c|cccccccc|c}
\toprule
\textbf{Methods} & \textbf{Ratio} & \textbf{CoLA} & \textbf{SST2} & \textbf{MRPC} & \textbf{STSB} & \textbf{QQP} & \textbf{MNLI} & \textbf{QNLI} & \textbf{RTE} & \textbf{Avg} \\
\midrule
{Individual} & 1$\times$ & 60.18  &  94.04  &  89.22  &  90.63  & 91.41  &  87.20  &  92.71  &  79.06 & 85.56   \\
\midrule
{BitDelta~\cite{liu2024bitdelta}} & 16$\times$ &  36.83  &  93.12  &  88.73  & 84.82 &  \textbf{90.00}  &  \textbf{85.57}   &  91.73   & 70.40 & 80.15  \\
{Delta-CoMe~\cite{ping2024delta}} & 16$\times$ &  57.67  &  92.48  &  87.99  & 90.53 &  89.17  &  82.66   &  \textbf{92.57}   & 76.9 & 83.74  \\
\midrule
{DARE~\cite{yu2024language} (Sparsity)} & 14.7$\times$ & 59.30 & 93.92 & 88.97 & 90.53 & 86.55 & 77.84 & 92.13 & 76.53 & 83.23 \\
{MP~\cite{han2015deep} (Sparsity)} & 14.7$\times$ & 57.69 & 92.43 & 84.31 & 88.18 & 86.26 & 74.40 & 86.80 & 70.40 & 80.06 \\
{DARE~\cite{yu2024language} (Compression)} & 31.7$\times$ & 58.22 & 93.69 & 87.50 & 90.27 & 53.47 & 36.48 & 89.51 & 75.81 & 73.12 \\
{MP~\cite{han2015deep} (Compression)} & 31.7$\times$ & 51.89 & 91.63 & 78.68 & 86.44 & 65.77 & 53.49 & 65.64 & 65.43 & 69.87 \\
\midrule
\textbf{\methodname{} (Ours)} & \textbf{32.9$\times$} &  \textbf{62.64}  &  \textbf{94.38}  &  \textbf{89.22}  & \textbf{90.60} &  85.79  &  82.92   &  92.15   & \textbf{77.98} & \textbf{84.46}  \\
\bottomrule
\end{tabular}
}
\end{table*}

%% file: tables/30_vit_b_16.tex
\begin{table*}[t]

\centering
\caption{Performance of \pruningshort{} and DARE on ViT-B/16 models across 30 tasks. ``(w/ $b$-bit)'' denotes the quantized setting using $b$-bit precision, while ``(w/o q.)'' refers to the non-quantized setting.}
\label{tab:30_vit} 
\resizebox{\textwidth}{!}{

\begin{tabular}{l|cccccccccc}
\toprule%

\textbf{Methods} & Animal-10N & Beans & Cats and Dogs & Cifar-10 & Cifar-100 & CUB-200-2011 & Dogs & DTD & EMNIST & EuroSAT\\ 


\midrule
Individual   & 92.46 & 97.70   & 99.05     & 97.88   & 89.85    & 84.78 & 89.91 & 81.1   & 94.67   & 99.07          \\

\midrule
DARE~\cite{liu2024bitdelta} (w/ 4-bit)  & 92.26 & 34.87 & 98.17 & 33.39 & 53.37 & 84.28 & 89.55 & 81.08 & 94.69 & 99.07 \\ 

\textbf{\methodname{} (w/ 4-bit) (Ours)} & 92.42 & 97.32 & 98.98 & 58.33 & 72.14 & 83.98 & 89.74 & 80.82 & 94.51 & 99.04 \\
\midrule

DARE~\cite{liu2024bitdelta} (w/ 8-bit) & 92.26 & 34.87 & 98.17 & 33.39 & 53.37 & 84.28 & 89.55 & 81.08 & 94.69 & 99.07 \\

\textbf{\methodname{} (w/ 8-bit) (Ours)} & 92.68 & 98.47 & 98.66 & 80.06 & 78.12 & 84.54 & 90.18 & 81.13 & 94.61 & 99.00 \\
\midrule

DARE~\cite{liu2024bitdelta} (w/o q.) & 92.28 & 97.70 & 98.04 & 57.90 & 56.07 & 84.19 & 89.91 & 81.15 & 94.58 & 99.15 \\

\textbf{\methodname{} (w/o q.) (Ours)} & 92.30 & 98.08 & 98.99 & 96.77 & 80.83 & 84.28 & 90.00 & 81.08 & 94.62 & 99.07 \\

\midrule
\midrule

& Fashion  & Flowers & KenyanFood13  & Food-101 & Fruits-360 & Garbage & GTSRB & Intel-Images & Kvasir-v2  & LandScape \\
\midrule
Individual  & 93.26 & 98.19 & 82.58 & 87.87 & 99.64 & 98.58 & 95.74 & 94.87 & 93.91 & 94.00 \\

\midrule
DARE~\cite{liu2024bitdelta} (w/ 4-bit) & 84.09 & 98.19 & 82.70 & 38.00 & 99.64 & 98.61 & 7.57 & 93.33 & 74.81 & 92.40 \\

\textbf{\methodname{} (w/ 4-bit) (Ours)}  & 79.58 & 98.17 & 82.21 & 29.14 & 99.58 & 98.54 & 87.16 & 90.93 & 56.47 & 93.80 \\

\midrule
DARE~\cite{liu2024bitdelta} (w/ 8-bit) & 84.09 & 98.19 & 82.70 & 38.00 & 99.64 & 98.61 & 75.70 & 93.33 & 74.81 & 92.40 \\

\textbf{\methodname{} (w/ 8-bit) (Ours)} & 83.85 & 98.19 & 82.45 & 47.03 & 99.59 & 98.50 & 91.96 & 93.93 & 89.94 & 73.80 \\

\midrule
DARE~\cite{liu2024bitdelta} (w/o q.) & 73.69 & 98.17 & 82.45 & 43.93 & 99.64 & 98.61 & 62.49 & 92.13 & 77.62 & 93.40 \\

\textbf{\methodname{} (w/o q.) (Ours)} & 90.10 & 98.12 & 82.82 & 61.87 & 99.63 & 98.50 & 95.36 & 94.13 & 92.22 & 94.20 \\
\midrule
\midrule

& MangoLeafBD & MNIST & Pet & RESISC45 & Cars & STL10 & SUN397 & SVHN & Vegetables & Weather \\
\midrule
Individual  & 100.00 & 99.22 & 92.20 & 99.00 & 85.29 & 99.08 & 87.50 & 96.22 & 100.00 & 98.19 \\

\midrule
DARE~\cite{liu2024bitdelta} (w/ 4-bit) & 98.47 & 78.59 & 91.88 & 98.90 & 79.78 & 99.09 & 85.24 & 96.22 & 91.67 & 97.62 \\

\textbf{\methodname{} (w/ 4-bit) (Ours)} & 85.95 & 98.93 & 92.56 & 98.95 & 79.74 & 99.21 & 85.82 & 96.21 & 91.30 & 80.46 \\

\midrule
DARE~\cite{liu2024bitdelta} (w/ 8-bit) & 98.47 & 78.59 & 91.88 & 98.90 & 79.78 & 99.09 & 85.24 & 96.22 & 91.67 & 97.62 \\

\textbf{\methodname{} (w/ 8-bit) (Ours)} & 100.00 & 98.68 & 92.37 & 99.10 & 80.11 & 99.16 & 85.74 & 96.33 & 99.67 & 97.51 \\

\midrule
DARE~\cite{liu2024bitdelta} (w/o q.) & 100.00 & 64.62 & 92.26 & 98.79 & 79.77 & 99.09 & 85.55 & 96.30 & 99.83 & 87.03 \\

\textbf{\methodname{} (w/o q.) (Ours)} & 100.00 & 98.97 & 92.64 & 99.13 & 80.55 & 99.22 & 85.90 & 96.32 & 99.97 & 97.90 \\

\midrule
\midrule

\textbf{Average Acc}& \multicolumn{1}{c|}{Individual}& \multicolumn{1}{|c|}{DARE (w/ 4-bit)}&
\multicolumn{2}{c|}{\textbf{\methodname{} (w/ 4-bit) (Ours)}}&
\multicolumn{1}{c|}{DARE (w/ 8-bit)}&
\multicolumn{2}{c|}{\textbf{\methodname{} (w/ 8-bit) (Ours)}}&
\multicolumn{1}{c|}{DARE (w/o q.)}&
\multicolumn{2}{c}{\textbf{\methodname{} (w/o q.) (Ours)}}
\\
\midrule
Acc& 
\multicolumn{1}{c|}{94.06}&
\multicolumn{1}{c|}{81.58}&
\multicolumn{2}{c|}{\textbf{86.40}}&
\multicolumn{1}{c|}{83.85}&
\multicolumn{2}{c|}{\textbf{90.18}}&
\multicolumn{1}{c|}{85.88}&
\multicolumn{2}{c}{\textbf{92.45}}

\\

\bottomrule
\end{tabular}

}
\end{table*}

%% file: tables/abl_hyperparameter.tex
\begin{table}[htbp]
\centering
\caption{Performance of different hyperparameter settings applied to ViT-B/32.}
\vspace{5pt}
\label{tab:ablation_hyperparameter}
\resizebox{0.99\textwidth}{!}{
\begin{tabular}{l|cccccccc|c}
\toprule 
\textbf{Method} & \textbf{SUN397} & \textbf{Cars} & \textbf{RESISC45} & \textbf{EuroSAT} & \textbf{SVHN} & \textbf{GTSRB} & \textbf{MNIST} & \textbf{DTD} & \textbf{Avg.} \\
\midrule
\multicolumn{10}{l}{\textbf{\pruningname{} (quantization bit)}} \\
\midrule
2-bit & 73.4 & 70.1 & 94.4 & 99.1 & 96.9 & 98.2 & 99.7 & 74.3 & 88.3 \\
4-bit & 78.6 & 77.4 & 95.7 & 99.4 & 97.0 & 98.7 & 99.7 & 79.0 & 90.7 \\
6-bit & 78.7 & 77.5 & 95.7 & 99.5 & 97.1 & 98.8 & 99.7 & 78.6 & 90.7 \\
8-bit & 78.8 & 77.7 & 95.7 & 99.7 & 97.0 & 98.7 & 99.7 & 79.0 & 90.8 \\
\midrule
\multicolumn{10}{l}{\textbf{\allocationname{} (sparsity step size)}} \\
\midrule
Step = 0.000 & 77.3 & 75.1 & 95.5 & 99.7 & 97.1 & 98.8 & 99.6 & 77.6 & 90.1 \\
Step = 0.005 & 77.9 & 75.7 & 95.4 & 99.7 & 97.0 & 98.8 & 99.7 & 78.9 & 90.4 \\
Step = 0.010 & 78.4 & 76.7 & 95.6 & 99.4 & 97.0 & 98.9 & 99.6 & 78.9 & 90.6 \\
Step = 0.020 & 78.6 & 77.4 & 95.7 & 99.4 & 97.0 & 98.7 & 99.7 & 79.0 & 90.7 \\
Step = 0.025 & 78.3 & 76.4 & 95.1 & 98.1 & 96.6 & 98.6 & 99.7 & 79.2 & 90.3 \\
Step = 0.030 & 78.5 & 75.3 & 94.5 & 96.5 & 96.7 & 98.2 & 99.7 & 78.7 & 89.8 \\
\bottomrule
\end{tabular}
}
\end{table}

%% file: tables/controlled_experiment.tex
\begin{table}[htbp]
\centering
\caption{Controlled experiments on ViT-B/32 with varying training steps. ``\methodname{} Accuracy'' denotes the test accuracy after applying \methodname{}.}
\vspace{5pt}
\label{tab:control_exp}
\resizebox{0.99\textwidth}{!}{
\begin{tabular}{l|cccccccccc}
\toprule 
\textbf{Training Steps} & \textbf{1000} & \textbf{2000} & \textbf{3000} & \textbf{4000} & \textbf{5000} & \textbf{6000} & \textbf{7000} & \textbf{8000} & \textbf{9000} & \textbf{10000} \\
\midrule
\multicolumn{11}{l}{\textbf{Cars~\cite{krause20133d}}} \\
\midrule
Training Loss        & 0.1404 & 0.0834 & 0.0206 & 0.0069 & 0.0059 & 0.0059 & 0.0034 & 0.0077 & 0.0048 & 0.0036 \\
Test Accuracy        & 72.86  & 68.90  & 75.22  & \textbf{78.78}  & \textbf{77.74}  & \textbf{78.37}  & \textbf{78.01}  & \textbf{75.21}  & \textbf{77.60}  & \textbf{77.34}  \\
\textbf{\methodname{} Accuracy} & \textbf{74.02}  & \textbf{73.27}  & \textbf{75.82}  & 75.56  & 75.62  & 75.66  & 76.13  & 72.80  & 74.74  & 74.06  \\
\midrule
\multicolumn{11}{l}{\textbf{SUN397~\cite{xiao2010sun}}} \\
\midrule
Training Loss        & 0.5475 & 0.1432 & 0.0426 & 0.1125 & 0.0526 & 0.0040 & 0.0018 & 0.0009 & 0.0006 & 0.0004 \\
Test Accuracy        & 73.63  & 75.94  & 75.40  & 74.63  & 74.07  & \textbf{76.94}  & 76.87  & 77.03  & 77.06  & 77.07  \\
\textbf{\methodname{} Accuracy} & \textbf{76.02}  & \textbf{76.92}  & \textbf{77.21}  & \textbf{76.20}  & \textbf{76.19}  & 76.66  & \textbf{76.94}  & \textbf{77.11}  & \textbf{77.19}  & \textbf{77.31}  \\
\bottomrule
\end{tabular}
}
\end{table}

%% file: neurips_2025.bbl
\begin{thebibliography}{93}
\providecommand{\natexlab}[1]{#1}
\providecommand{\url}[1]{\texttt{#1}}
\expandafter\ifx\csname urlstyle\endcsname\relax
  \providecommand{\doi}[1]{doi: #1}\else
  \providecommand{\doi}{doi: \begingroup \urlstyle{rm}\Url}\fi

\bibitem[Ahmed et~al.(2021)Ahmed, Mamun, and Asif]{ahmed2021dcnn}
M~Israk Ahmed, Shahriyar~Mahmud Mamun, and Asif Uz~Zaman Asif.
\newblock Dcnn-based vegetable image classification using transfer learning: A comparative study.
\newblock In \emph{2021 5th International Conference on Computer, Communication and Signal Processing (ICCCSP)}, pages 235--243. IEEE, 2021.

\bibitem[Ahmed et~al.(2023)Ahmed, Ibrahim, Nadim, Rahman, Shejunti, Jabid, and Ali]{ahmed2023mangoleafbd}
Sarder~Iftekhar Ahmed, Muhammad Ibrahim, Md Nadim, Md~Mizanur Rahman, Maria~Mehjabin Shejunti, Taskeed Jabid, and Md~Sawkat Ali.
\newblock Mangoleafbd: A comprehensive image dataset to classify diseased and healthy mango leaves.
\newblock \emph{Data in Brief}, 47:\penalty0 108941, 2023.

\bibitem[Austin et~al.(2021)Austin, Odena, Nye, Bosma, Michalewski, Dohan, Jiang, Cai, Terry, Le, et~al.]{austin2021program}
Jacob Austin, Augustus Odena, Maxwell Nye, Maarten Bosma, Henryk Michalewski, David Dohan, Ellen Jiang, Carrie Cai, Michael Terry, Quoc Le, et~al.
\newblock Program synthesis with large language models.
\newblock \emph{arXiv preprint arXiv:2108.07732}, 2021.

\bibitem[Banerjee and Lavie(2005)]{banerjee2005meteor}
Satanjeev Banerjee and Alon Lavie.
\newblock Meteor: An automatic metric for mt evaluation with improved correlation with human judgments.
\newblock In \emph{Proceedings of the acl workshop on intrinsic and extrinsic evaluation measures for machine translation and/or summarization}, pages 65--72, 2005.

\bibitem[Bansal(2019)]{bansal2019intel}
Puneet Bansal.
\newblock Intel image classification.
\newblock \emph{Available on https://www. kaggle. com/puneet6060/intel-image-classification, Online}, 2019.

\bibitem[Berger(2003)]{berger2003rate}
Toby Berger.
\newblock Rate-distortion theory.
\newblock \emph{Wiley Encyclopedia of Telecommunications}, 2003.

\bibitem[Bossard et~al.(2014)Bossard, Guillaumin, and Van~Gool]{bossard14}
Lukas Bossard, Matthieu Guillaumin, and Luc Van~Gool.
\newblock Food-101 -- mining discriminative components with random forests.
\newblock In \emph{European Conference on Computer Vision}, 2014.

\bibitem[Caruana(1997)]{caruana1997multitask}
Rich Caruana.
\newblock Multitask learning.
\newblock \emph{Machine learning}, 28:\penalty0 41--75, 1997.

\bibitem[CCHANG(2018)]{cchang_2018}
CCHANG.
\newblock Garbage classification.
\newblock \url{https://www.kaggle.com/ds/81794}, 2018.

\bibitem[Cer et~al.(2017)Cer, Diab, Agirre, Lopez-Gazpio, and Specia]{cer2017semeval}
Daniel Cer, Mona Diab, Eneko Agirre, Inigo Lopez-Gazpio, and Lucia Specia.
\newblock Semeval-2017 task 1: Semantic textual similarity-multilingual and cross-lingual focused evaluation.
\newblock \emph{arXiv preprint arXiv:1708.00055}, 2017.

\bibitem[Chen et~al.(2021)Chen, Tworek, Jun, Yuan, Pinto, Kaplan, Edwards, Burda, Joseph, Brockman, et~al.]{chen2021evaluating}
Mark Chen, Jerry Tworek, Heewoo Jun, Qiming Yuan, Henrique Ponde De~Oliveira Pinto, Jared Kaplan, Harri Edwards, Yuri Burda, Nicholas Joseph, Greg Brockman, et~al.
\newblock Evaluating large language models trained on code.
\newblock \emph{arXiv preprint arXiv:2107.03374}, 2021.

\bibitem[Cheng et~al.(2017)Cheng, Han, and Lu]{cheng2017remote}
Gong Cheng, Junwei Han, and Xiaoqiang Lu.
\newblock Remote sensing image scene classification: Benchmark and state of the art.
\newblock \emph{Proceedings of the IEEE}, 105\penalty0 (10):\penalty0 1865--1883, 2017.

\bibitem[Cimpoi et~al.(2014)Cimpoi, Maji, Kokkinos, Mohamed, and Vedaldi]{cimpoi2014describing}
Mircea Cimpoi, Subhransu Maji, Iasonas Kokkinos, Sammy Mohamed, and Andrea Vedaldi.
\newblock Describing textures in the wild.
\newblock In \emph{Proceedings of the IEEE conference on computer vision and pattern recognition}, pages 3606--3613, 2014.

\bibitem[Coates et~al.(2011)Coates, Ng, and Lee]{coates2011analysis}
Adam Coates, Andrew Ng, and Honglak Lee.
\newblock An analysis of single-layer networks in unsupervised feature learning.
\newblock In \emph{Proceedings of the fourteenth international conference on artificial intelligence and statistics}, pages 215--223. JMLR Workshop and Conference Proceedings, 2011.

\bibitem[Cobbe et~al.(2021)Cobbe, Kosaraju, Bavarian, Chen, Jun, Kaiser, Plappert, Tworek, Hilton, Nakano, et~al.]{cobbe2021training}
Karl Cobbe, Vineet Kosaraju, Mohammad Bavarian, Mark Chen, Heewoo Jun, Lukasz Kaiser, Matthias Plappert, Jerry Tworek, Jacob Hilton, Reiichiro Nakano, et~al.
\newblock Training verifiers to solve math word problems.
\newblock \emph{arXiv preprint arXiv:2110.14168}, 2021.

\bibitem[Cohen et~al.(2017)Cohen, Afshar, Tapson, and Van~Schaik]{cohen2017emnist}
Gregory Cohen, Saeed Afshar, Jonathan Tapson, and Andre Van~Schaik.
\newblock Emnist: Extending mnist to handwritten letters.
\newblock In \emph{2017 international joint conference on neural networks (IJCNN)}, pages 2921--2926. IEEE, 2017.

\bibitem[Cover(1999)]{cover1999elements}
Thomas~M Cover.
\newblock \emph{Elements of information theory}.
\newblock John Wiley \& Sons, 1999.

\bibitem[Cukierski()]{cukierskidogs}
Will Cukierski.
\newblock Dogs vs. cats, 2013.
\newblock \emph{URL https://kaggle. com/competitions/dogs-vs-cats}.

\bibitem[DeepNets()]{Landscape}
DeepNets.
\newblock Landscape recognition.
\newblock \url{https://www.kaggle.com/datasets/utkarshsaxenadn/landscape-recognition-image-dataset-12k-images}.

\bibitem[Deng et~al.(2024)Deng, Zhao, Vakilian, Chen, Li, and Thrampoulidis]{deng2024dare}
Wenlong Deng, Yize Zhao, Vala Vakilian, Minghui Chen, Xiaoxiao Li, and Christos Thrampoulidis.
\newblock Dare the extreme: Revisiting delta-parameter pruning for fine-tuned models.
\newblock \emph{arXiv preprint arXiv:2410.09344}, 2024.

\bibitem[Dolan and Brockett(2005)]{dolan2005automatically}
Bill Dolan and Chris Brockett.
\newblock Automatically constructing a corpus of sentential paraphrases.
\newblock In \emph{Third international workshop on paraphrasing (IWP2005)}, 2005.

\bibitem[Frantar et~al.(2022)Frantar, Ashkboos, Hoefler, and Alistarh]{frantar2022gptq}
Elias Frantar, Saleh Ashkboos, Torsten Hoefler, and Dan Alistarh.
\newblock Gptq: Accurate post-training quantization for generative pre-trained transformers.
\newblock \emph{arXiv preprint arXiv:2210.17323}, 2022.

\bibitem[Gao et~al.(2024)Gao, Tow, Abbasi, Biderman, Black, DiPofi, Foster, Golding, Hsu, Le~Noac'h, Li, McDonell, Muennighoff, Ociepa, Phang, Reynolds, Schoelkopf, Skowron, Sutawika, Tang, Thite, Wang, Wang, and Zou]{eval-harness}
Leo Gao, Jonathan Tow, Baber Abbasi, Stella Biderman, Sid Black, Anthony DiPofi, Charles Foster, Laurence Golding, Jeffrey Hsu, Alain Le~Noac'h, Haonan Li, Kyle McDonell, Niklas Muennighoff, Chris Ociepa, Jason Phang, Laria Reynolds, Hailey Schoelkopf, Aviya Skowron, Lintang Sutawika, Eric Tang, Anish Thite, Ben Wang, Kevin Wang, and Andy Zou.
\newblock The language model evaluation harness, 2024.

\bibitem[Giampiccolo et~al.(2007)Giampiccolo, Magnini, Dagan, and Dolan]{giampiccolo2007third}
Danilo Giampiccolo, Bernardo Magnini, Ido Dagan, and William~B Dolan.
\newblock The third pascal recognizing textual entailment challenge.
\newblock In \emph{Proceedings of the ACL-PASCAL workshop on textual entailment and paraphrasing}, pages 1--9, 2007.

\bibitem[Golomb(1966)]{golomb1966run}
Solomon Golomb.
\newblock Run-length encodings (corresp.).
\newblock \emph{IEEE transactions on information theory}, 12\penalty0 (3):\penalty0 399--401, 1966.

\bibitem[Goyal et~al.(2017)Goyal, Khot, Summers-Stay, Batra, and Parikh]{goyal2017making}
Yash Goyal, Tejas Khot, Douglas Summers-Stay, Dhruv Batra, and Devi Parikh.
\newblock Making the v in vqa matter: Elevating the role of image understanding in visual question answering.
\newblock In \emph{Proceedings of the IEEE conference on computer vision and pattern recognition}, pages 6904--6913, 2017.

\bibitem[Han et~al.(2015{\natexlab{a}})Han, Mao, and Dally]{han2015deep}
Song Han, Huizi Mao, and William~J Dally.
\newblock Deep compression: Compressing deep neural networks with pruning, trained quantization and huffman coding.
\newblock \emph{arXiv preprint arXiv:1510.00149}, 2015{\natexlab{a}}.

\bibitem[Han et~al.(2015{\natexlab{b}})Han, Pool, Tran, and Dally]{han2015learning}
Song Han, Jeff Pool, John Tran, and William Dally.
\newblock Learning both weights and connections for efficient neural network.
\newblock \emph{Advances in neural information processing systems}, 28, 2015{\natexlab{b}}.

\bibitem[Helber et~al.(2019)Helber, Bischke, Dengel, and Borth]{helber2019eurosat}
Patrick Helber, Benjamin Bischke, Andreas Dengel, and Damian Borth.
\newblock Eurosat: A novel dataset and deep learning benchmark for land use and land cover classification.
\newblock \emph{IEEE Journal of Selected Topics in Applied Earth Observations and Remote Sensing}, 12\penalty0 (7):\penalty0 2217--2226, 2019.

\bibitem[Hendrycks et~al.(2021{\natexlab{a}})Hendrycks, Burns, Basart, Zou, Mazeika, Song, and Steinhardt]{hendryckstest2021}
Dan Hendrycks, Collin Burns, Steven Basart, Andy Zou, Mantas Mazeika, Dawn Song, and Jacob Steinhardt.
\newblock Measuring massive multitask language understanding.
\newblock \emph{Proceedings of the International Conference on Learning Representations (ICLR)}, 2021{\natexlab{a}}.

\bibitem[Hendrycks et~al.(2021{\natexlab{b}})Hendrycks, Burns, Kadavath, Arora, Basart, Tang, Song, and Steinhardt]{hendrycks2021measuring}
Dan Hendrycks, Collin Burns, Saurav Kadavath, Akul Arora, Steven Basart, Eric Tang, Dawn Song, and Jacob Steinhardt.
\newblock Measuring mathematical problem solving with the {MATH} dataset.
\newblock In \emph{Thirty-fifth Conference on Neural Information Processing Systems Datasets and Benchmarks Track (Round 2)}, 2021{\natexlab{b}}.

\bibitem[Huang et~al.(2024)Huang, Ye, Chen, He, Yue, and Ouyang]{huang2024emr}
Chenyu Huang, Peng Ye, Tao Chen, Tong He, Xiangyu Yue, and Wanli Ouyang.
\newblock Emr-merging: Tuning-free high-performance model merging.
\newblock \emph{Advances in Neural Information Processing Systems}, 37:\penalty0 122741--122769, 2024.

\bibitem[Huang et~al.(2025)Huang, Ye, Wang, Zheng, Qi, Bai, Ouyang, and Chen]{huang2025seeing}
Chenyu Huang, Peng Ye, Xiaohui Wang, Shenghe Zheng, Biqing Qi, Lei Bai, Wanli Ouyang, and Tao Chen.
\newblock Seeing delta parameters as jpeg images: Data-free delta compression with discrete cosine transform.
\newblock \emph{arXiv preprint arXiv:2503.06676}, 2025.

\bibitem[Ilharco et~al.(2022)Ilharco, Ribeiro, Wortsman, Gururangan, Schmidt, Hajishirzi, and Farhadi]{ilharco2022editing}
Gabriel Ilharco, Marco~Tulio Ribeiro, Mitchell Wortsman, Suchin Gururangan, Ludwig Schmidt, Hannaneh Hajishirzi, and Ali Farhadi.
\newblock Editing models with task arithmetic.
\newblock \emph{arXiv preprint arXiv:2212.04089}, 2022.

\bibitem[Isik et~al.(2023)Isik, Kumbong, Ning, Yao, Koyejo, and Zhang]{isik2023gpt}
Berivan Isik, Hermann Kumbong, Wanyi Ning, Xiaozhe Yao, Sanmi Koyejo, and Ce Zhang.
\newblock Gpt-zip: Deep compression of finetuned large language models.
\newblock In \emph{Workshop on Efficient Systems for Foundation Models@ ICML2023}, 2023.

\bibitem[Iyer et~al.(2017)Iyer, Dandekar, Csernai, et~al.]{iyer2017first}
Shankar Iyer, Nikhil Dandekar, Korn{\'e}l Csernai, et~al.
\newblock First quora dataset release: Question pairs. data. quora. com.
\newblock 2017.

\bibitem[Jalal et~al.(2019)Jalal, Wang, Jefferson, Zheng, Nsoesie, and Betke]{jalal2019scraping}
Mona Jalal, Kaihong Wang, Sankara Jefferson, Yi Zheng, Elaine~O Nsoesie, and Margrit Betke.
\newblock Scraping social media photos posted in kenya and elsewhere to detect and analyze food types.
\newblock In \emph{Proceedings of the 5th International Workshop on Multimedia Assisted Dietary Management}, pages 50--59, 2019.

\bibitem[Jiang et~al.(2024)Jiang, Yang, Chen, Li, Li, and Li]{jiang2024deltadq}
Yanfeng Jiang, Zelan Yang, Bohua Chen, Shen Li, Yong Li, and Tao Li.
\newblock Deltadq: Ultra-high delta compression for fine-tuned llms via group-wise dropout and separate quantization.
\newblock \emph{arXiv preprint arXiv:2410.08666}, 2024.

\bibitem[Khosla et~al.(2011)Khosla, Jayadevaprakash, Yao, and Fei-Fei]{KhoslaYaoJayadevaprakashFeiFei_FGVC2011}
Aditya Khosla, Nityananda Jayadevaprakash, Bangpeng Yao, and Li Fei-Fei.
\newblock Novel dataset for fine-grained image categorization.
\newblock In \emph{First Workshop on Fine-Grained Visual Categorization, IEEE Conference on Computer Vision and Pattern Recognition}, Colorado Springs, CO, 2011.

\bibitem[Krause et~al.(2013)Krause, Stark, Deng, and Fei-Fei]{krause20133d}
Jonathan Krause, Michael Stark, Jia Deng, and Li Fei-Fei.
\newblock 3d object representations for fine-grained categorization.
\newblock In \emph{Proceedings of the IEEE international conference on computer vision workshops}, pages 554--561, 2013.

\bibitem[Krizhevsky et~al.(2009)Krizhevsky, Hinton, et~al.]{krizhevsky2009learning}
Alex Krizhevsky, Geoffrey Hinton, et~al.
\newblock Learning multiple layers of features from tiny images.
\newblock 2009.

\bibitem[Lab(2020)]{beansdata}
Makerere~AI Lab.
\newblock Bean disease dataset, 2020.

\bibitem[Lambert et~al.(2024)Lambert, Morrison, Pyatkin, Huang, Ivison, Brahman, Miranda, Liu, Dziri, Lyu, Gu, Malik, Graf, Hwang, Yang, Bras, Tafjord, Wilhelm, Soldaini, Smith, Wang, Dasigi, and Hajishirzi]{lambert2024tulu3}
Nathan Lambert, Jacob Morrison, Valentina Pyatkin, Shengyi Huang, Hamish Ivison, Faeze Brahman, Lester James~V. Miranda, Alisa Liu, Nouha Dziri, Shane Lyu, Yuling Gu, Saumya Malik, Victoria Graf, Jena~D. Hwang, Jiangjiang Yang, Ronan~Le Bras, Oyvind Tafjord, Chris Wilhelm, Luca Soldaini, Noah~A. Smith, Yizhong Wang, Pradeep Dasigi, and Hannaneh Hajishirzi.
\newblock Tülu 3: Pushing frontiers in open language model post-training.
\newblock 2024.

\bibitem[LeCun(1998)]{lecun1998mnist}
Yann LeCun.
\newblock The mnist database of handwritten digits.
\newblock \emph{http://yann. lecun. com/exdb/mnist/}, 1998.

\bibitem[Lee et~al.(2020)Lee, Park, Mo, Ahn, and Shin]{lee2020layer}
Jaeho Lee, Sejun Park, Sangwoo Mo, Sungsoo Ahn, and Jinwoo Shin.
\newblock Layer-adaptive sparsity for the magnitude-based pruning.
\newblock \emph{arXiv preprint arXiv:2010.07611}, 2020.

\bibitem[Li et~al.(2018)Li, Qian, Jiang, Lu, and Tang]{li2018optimization}
Guiying Li, Chao Qian, Chunhui Jiang, Xiaofen Lu, and Ke Tang.
\newblock Optimization based layer-wise magnitude-based pruning for dnn compression.
\newblock In \emph{IJCAI}, pages 2383--2389, 2018.

\bibitem[Li et~al.(2024)Li, Li, Lee, and Sun]{li2024adaptive}
Wei Li, Lujun Li, Mark Lee, and Shengjie Sun.
\newblock Adaptive layer sparsity for large language models via activation correlation assessment.
\newblock \emph{Advances in Neural Information Processing Systems}, 37:\penalty0 109350--109380, 2024.

\bibitem[Lin(2004)]{lin2004rouge}
Chin-Yew Lin.
\newblock Rouge: A package for automatic evaluation of summaries.
\newblock In \emph{Text summarization branches out}, pages 74--81, 2004.

\bibitem[Lin et~al.(2021)Lin, Hilton, and Evans]{lin2021truthfulqa}
Stephanie Lin, Jacob Hilton, and Owain Evans.
\newblock Truthfulqa: Measuring how models mimic human falsehoods.
\newblock \emph{arXiv preprint arXiv:2109.07958}, 2021.

\bibitem[Lin et~al.(2024)Lin, Lyu, Liu, Tang, Liang, Song, and Chang]{lin2024mlp}
Sihao Lin, Pumeng Lyu, Dongrui Liu, Tao Tang, Xiaodan Liang, Andy Song, and Xiaojun Chang.
\newblock Mlp can be a good transformer learner.
\newblock In \emph{Proceedings of the IEEE/CVF Conference on Computer Vision and Pattern Recognition}, pages 19489--19498, 2024.

\bibitem[Lin et~al.(2014)Lin, Maire, Belongie, Hays, Perona, Ramanan, Doll{\'a}r, and Zitnick]{lin2014microsoft}
Tsung-Yi Lin, Michael Maire, Serge Belongie, James Hays, Pietro Perona, Deva Ramanan, Piotr Doll{\'a}r, and C~Lawrence Zitnick.
\newblock Microsoft coco: Common objects in context.
\newblock In \emph{Computer Vision--ECCV 2014: 13th European Conference, Zurich, Switzerland, September 6-12, 2014, Proceedings, Part V 13}, pages 740--755. Springer, 2014.

\bibitem[Liu et~al.(2023)Liu, Xia, Wang, and Zhang]{evalplus}
Jiawei Liu, Chunqiu~Steven Xia, Yuyao Wang, and Lingming Zhang.
\newblock Is your code generated by chat{GPT} really correct? rigorous evaluation of large language models for code generation.
\newblock In \emph{Thirty-seventh Conference on Neural Information Processing Systems}, 2023.

\bibitem[Liu et~al.(2024)Liu, Xiao, Li, Lee, Han, Dao, and Cai]{liu2024bitdelta}
James Liu, Guangxuan Xiao, Kai Li, Jason~D Lee, Song Han, Tri Dao, and Tianle Cai.
\newblock Bitdelta: Your fine-tune may only be worth one bit.
\newblock \emph{Advances in Neural Information Processing Systems}, 37:\penalty0 13579--13600, 2024.

\bibitem[Liu et~al.(2019)Liu, Ott, Goyal, Du, Joshi, Chen, Levy, Lewis, Zettlemoyer, and Stoyanov]{liu2019roberta}
Yinhan Liu, Myle Ott, Naman Goyal, Jingfei Du, Mandar Joshi, Danqi Chen, Omer Levy, Mike Lewis, Luke Zettlemoyer, and Veselin Stoyanov.
\newblock Roberta: A robustly optimized bert pretraining approach.
\newblock \emph{arXiv preprint arXiv:1907.11692}, 2019.

\bibitem[Lu et~al.(2024)Lu, Zhou, Liu, Wang, Mahoney, and Yang]{lu2024alphapruning}
Haiquan Lu, Yefan Zhou, Shiwei Liu, Zhangyang Wang, Michael~W Mahoney, and Yaoqing Yang.
\newblock Alphapruning: Using heavy-tailed self regularization theory for improved layer-wise pruning of large language models.
\newblock \emph{Advances in neural information processing systems}, 37:\penalty0 9117--9152, 2024.

\bibitem[Luo et~al.(2023{\natexlab{a}})Luo, Sun, Xu, Zhao, Lou, Tao, Geng, Lin, Chen, and Zhang]{luo2023wizardmath}
Haipeng Luo, Qingfeng Sun, Can Xu, Pu Zhao, Jianguang Lou, Chongyang Tao, Xiubo Geng, Qingwei Lin, Shifeng Chen, and Dongmei Zhang.
\newblock Wizardmath: Empowering mathematical reasoning for large language models via reinforced evol-instruct.
\newblock \emph{arXiv preprint arXiv:2308.09583}, 2023{\natexlab{a}}.

\bibitem[Luo et~al.(2023{\natexlab{b}})Luo, Xu, Zhao, Sun, Geng, Hu, Tao, Ma, Lin, and Jiang]{luo2023wizardcoder}
Ziyang Luo, Can Xu, Pu Zhao, Qingfeng Sun, Xiubo Geng, Wenxiang Hu, Chongyang Tao, Jing Ma, Qingwei Lin, and Daxin Jiang.
\newblock Wizardcoder: Empowering code large language models with evol-instruct.
\newblock \emph{arXiv preprint arXiv:2306.08568}, 2023{\natexlab{b}}.

\bibitem[Mamaev()]{Flowers}
Alexander Mamaev.
\newblock Flowers recognition.
\newblock \url{https://www.kaggle.com/datasets/alxmamaev/flowers-recognition}.

\bibitem[Muresan and Oltean(2018)]{muresan2018fruit}
Horea Muresan and Mihai Oltean.
\newblock Fruit recognition from images using deep learning.
\newblock \emph{Acta Universitatis Sapientiae, Informatica}, 10\penalty0 (1):\penalty0 26--42, 2018.

\bibitem[Netzer et~al.(2011)Netzer, Wang, Coates, Bissacco, Wu, Ng, et~al.]{netzer2011reading}
Yuval Netzer, Tao Wang, Adam Coates, Alessandro Bissacco, Baolin Wu, Andrew~Y Ng, et~al.
\newblock Reading digits in natural images with unsupervised feature learning.
\newblock In \emph{NIPS workshop on deep learning and unsupervised feature learning}, page~4. Granada, 2011.

\bibitem[Papineni et~al.(2002)Papineni, Roukos, Ward, and Zhu]{papineni2002bleu}
Kishore Papineni, Salim Roukos, Todd Ward, and Wei-Jing Zhu.
\newblock Bleu: a method for automatic evaluation of machine translation.
\newblock In \emph{Proceedings of the 40th annual meeting of the Association for Computational Linguistics}, pages 311--318, 2002.

\bibitem[Parkhi et~al.(2012)Parkhi, Vedaldi, Zisserman, and Jawahar]{parkhi2012cats}
Omkar~M Parkhi, Andrea Vedaldi, Andrew Zisserman, and CV Jawahar.
\newblock Cats and dogs.
\newblock In \emph{2012 IEEE conference on computer vision and pattern recognition}, pages 3498--3505. IEEE, 2012.

\bibitem[Parrish et~al.(2021)Parrish, Chen, Nangia, Padmakumar, Phang, Thompson, Htut, and Bowman]{parrish2021bbq}
Alicia Parrish, Angelica Chen, Nikita Nangia, Vishakh Padmakumar, Jason Phang, Jana Thompson, Phu~Mon Htut, and Samuel~R Bowman.
\newblock Bbq: A hand-built bias benchmark for question answering.
\newblock \emph{arXiv preprint arXiv:2110.08193}, 2021.

\bibitem[Ping et~al.(2024)Ping, Wang, Wang, Han, Xu, Yan, Chen, Chang, Liu, and Sun]{ping2024delta}
Bowen Ping, Shuo Wang, Hanqing Wang, Xu Han, Yuzhuang Xu, Yukun Yan, Yun Chen, Baobao Chang, Zhiyuan Liu, and Maosong Sun.
\newblock Delta-come: Training-free delta-compression with mixed-precision for large language models.
\newblock \emph{Advances in Neural Information Processing Systems}, 37:\penalty0 31056--31077, 2024.

\bibitem[Pogorelov et~al.(2017)Pogorelov, Randel, Griwodz, Eskeland, de~Lange, Johansen, Spampinato, Dang-Nguyen, Lux, Schmidt, et~al.]{pogorelov2017kvasir}
Konstantin Pogorelov, Kristin~Ranheim Randel, Carsten Griwodz, Sigrun~Losada Eskeland, Thomas de Lange, Dag Johansen, Concetto Spampinato, Duc-Tien Dang-Nguyen, Mathias Lux, Peter~Thelin Schmidt, et~al.
\newblock Kvasir: A multi-class image dataset for computer aided gastrointestinal disease detection.
\newblock In \emph{Proceedings of the 8th ACM on Multimedia Systems Conference}, pages 164--169, 2017.

\bibitem[Radford et~al.(2021)Radford, Kim, Hallacy, Ramesh, Goh, Agarwal, Sastry, Askell, Mishkin, Clark, et~al.]{radford2021learning}
Alec Radford, Jong~Wook Kim, Chris Hallacy, Aditya Ramesh, Gabriel Goh, Sandhini Agarwal, Girish Sastry, Amanda Askell, Pamela Mishkin, Jack Clark, et~al.
\newblock Learning transferable visual models from natural language supervision.
\newblock In \emph{International conference on machine learning}, pages 8748--8763. PmLR, 2021.

\bibitem[Raffel et~al.(2020)Raffel, Shazeer, Roberts, Lee, Narang, Matena, Zhou, Li, and Liu]{raffel2020exploring}
Colin Raffel, Noam Shazeer, Adam Roberts, Katherine Lee, Sharan Narang, Michael Matena, Yanqi Zhou, Wei Li, and Peter~J Liu.
\newblock Exploring the limits of transfer learning with a unified text-to-text transformer.
\newblock \emph{Journal of machine learning research}, 21\penalty0 (140):\penalty0 1--67, 2020.

\bibitem[Rajpurkar et~al.(2016)Rajpurkar, Zhang, Lopyrev, and Liang]{rajpurkar2016squad}
Pranav Rajpurkar, Jian Zhang, Konstantin Lopyrev, and Percy Liang.
\newblock Squad: 100,000+ questions for machine comprehension of text.
\newblock \emph{arXiv preprint arXiv:1606.05250}, 2016.

\bibitem[Rein et~al.(2024)Rein, Hou, Stickland, Petty, Pang, Dirani, Michael, and Bowman]{rein2024gpqa}
David Rein, Betty~Li Hou, Asa~Cooper Stickland, Jackson Petty, Richard~Yuanzhe Pang, Julien Dirani, Julian Michael, and Samuel~R Bowman.
\newblock Gpqa: A graduate-level google-proof q\&a benchmark.
\newblock In \emph{First Conference on Language Modeling}, 2024.

\bibitem[Sattler et~al.(2019)Sattler, Wiedemann, M{\"u}ller, and Samek]{sattler2019sparse}
Felix Sattler, Simon Wiedemann, Klaus-Robert M{\"u}ller, and Wojciech Samek.
\newblock Sparse binary compression: Towards distributed deep learning with minimal communication.
\newblock In \emph{2019 International Joint Conference on Neural Networks (IJCNN)}, pages 1--8. IEEE, 2019.

\bibitem[Socher et~al.(2013)Socher, Perelygin, Wu, Chuang, Manning, Ng, and Potts]{socher2013recursive}
Richard Socher, Alex Perelygin, Jean Wu, Jason Chuang, Christopher~D Manning, Andrew~Y Ng, and Christopher Potts.
\newblock Recursive deep models for semantic compositionality over a sentiment treebank.
\newblock In \emph{Proceedings of the 2013 conference on empirical methods in natural language processing}, pages 1631--1642, 2013.

\bibitem[Song et~al.(2019)Song, Kim, and Lee]{song2019selfie}
Hwanjun Song, Minseok Kim, and Jae-Gil Lee.
\newblock Selfie: Refurbishing unclean samples for robust deep learning.
\newblock In \emph{International conference on machine learning}, pages 5907--5915. PMLR, 2019.

\bibitem[Stallkamp et~al.(2011)Stallkamp, Schlipsing, Salmen, and Igel]{stallkamp2011german}
Johannes Stallkamp, Marc Schlipsing, Jan Salmen, and Christian Igel.
\newblock The german traffic sign recognition benchmark: a multi-class classification competition.
\newblock In \emph{The 2011 international joint conference on neural networks}, pages 1453--1460. IEEE, 2011.

\bibitem[Strom(2015)]{strom15_interspeech}
Nikko Strom.
\newblock Scalable distributed dnn training using commodity gpu cloud computing.
\newblock In \emph{Interspeech 2015}, pages 1488--1492, 2015.

\bibitem[Suhr et~al.(2018)Suhr, Zhou, Zhang, Zhang, Bai, and Artzi]{suhr2018corpus}
Alane Suhr, Stephanie Zhou, Ally Zhang, Iris Zhang, Huajun Bai, and Yoav Artzi.
\newblock A corpus for reasoning about natural language grounded in photographs.
\newblock \emph{arXiv preprint arXiv:1811.00491}, 2018.

\bibitem[Tang et~al.(2024)Tang, Shen, Luo, Hu, Du, and Tao]{tang2024fusionbench}
Anke Tang, Li Shen, Yong Luo, Han Hu, Bo Du, and Dacheng Tao.
\newblock Fusionbench: A comprehensive benchmark of deep model fusion.
\newblock \emph{arXiv preprint arXiv:2406.03280}, 2024.

\bibitem[Team(2025)]{qwen3guard}
Qwen Team.
\newblock Qwen3guard technical report.
\newblock 2025.

\bibitem[Touvron et~al.(2023)Touvron, Martin, Stone, Albert, Almahairi, Babaei, Bashlykov, Batra, Bhargava, Bhosale, et~al.]{touvron2023llama}
Hugo Touvron, Louis Martin, Kevin Stone, Peter Albert, Amjad Almahairi, Yasmine Babaei, Nikolay Bashlykov, Soumya Batra, Prajjwal Bhargava, Shruti Bhosale, et~al.
\newblock Llama 2: Open foundation and fine-tuned chat models.
\newblock \emph{arXiv preprint arXiv:2307.09288}, 2023.

\bibitem[Vedantam et~al.(2015)Vedantam, Lawrence~Zitnick, and Parikh]{vedantam2015cider}
Ramakrishna Vedantam, C Lawrence~Zitnick, and Devi Parikh.
\newblock Cider: Consensus-based image description evaluation.
\newblock In \emph{Proceedings of the IEEE conference on computer vision and pattern recognition}, pages 4566--4575, 2015.

\bibitem[Wah et~al.(2011)Wah, Branson, Welinder, Perona, and Belongie]{wah2011caltech}
Catherine Wah, Steve Branson, Peter Welinder, Pietro Perona, and Serge Belongie.
\newblock The caltech-ucsd birds-200-2011 dataset.
\newblock 2011.

\bibitem[Wang et~al.(2018)Wang, Singh, Michael, Hill, Levy, and Bowman]{wang2018glue}
Alex Wang, Amanpreet Singh, Julian Michael, Felix Hill, Omer Levy, and Samuel~R Bowman.
\newblock Glue: A multi-task benchmark and analysis platform for natural language understanding.
\newblock \emph{arXiv preprint arXiv:1804.07461}, 2018.

\bibitem[Wang et~al.(2023)Wang, Bao, Dong, Bjorck, Peng, Liu, Aggarwal, Mohammed, Singhal, Som, and Wei]{beit3}
Wenhui Wang, Hangbo Bao, Li Dong, Johan Bjorck, Zhiliang Peng, Qiang Liu, Kriti Aggarwal, Owais~Khan Mohammed, Saksham Singhal, Subhojit Som, and Furu Wei.
\newblock Image as a foreign language: {BEiT} pretraining for vision and vision-language tasks.
\newblock In \emph{Proceedings of the IEEE/CVF Conference on Computer Vision and Pattern Recognition}, 2023.

\bibitem[Warstadt et~al.(2019)Warstadt, Singh, and Bowman]{warstadt2019neural}
Alex Warstadt, Amanpreet Singh, and Samuel~R Bowman.
\newblock Neural network acceptability judgments.
\newblock \emph{Transactions of the Association for Computational Linguistics}, 7:\penalty0 625--641, 2019.

\bibitem[Williams et~al.(2017)Williams, Nangia, and Bowman]{williams2017broad}
Adina Williams, Nikita Nangia, and Samuel~R Bowman.
\newblock A broad-coverage challenge corpus for sentence understanding through inference.
\newblock \emph{arXiv preprint arXiv:1704.05426}, 2017.

\bibitem[Xiao et~al.(2017)Xiao, Rasul, and Vollgraf]{xiao2017fashion}
Han Xiao, Kashif Rasul, and Roland Vollgraf.
\newblock Fashion-mnist: a novel image dataset for benchmarking machine learning algorithms.
\newblock \emph{arXiv preprint arXiv:1708.07747}, 2017.

\bibitem[Xiao et~al.(2021)Xiao, Zhang, Shen, Wu, and Zhang]{xiao2021classification}
Haixia Xiao, Feng Zhang, Zhongping Shen, Kun Wu, and Jinglin Zhang.
\newblock Classification of weather phenomenon from images by using deep convolutional neural network.
\newblock \emph{Earth and Space Science}, 8\penalty0 (5):\penalty0 e2020EA001604, 2021.

\bibitem[Xiao et~al.(2010)Xiao, Hays, Ehinger, Oliva, and Torralba]{xiao2010sun}
Jianxiong Xiao, James Hays, Krista~A Ehinger, Aude Oliva, and Antonio Torralba.
\newblock Sun database: Large-scale scene recognition from abbey to zoo.
\newblock In \emph{2010 IEEE computer society conference on computer vision and pattern recognition}, pages 3485--3492. IEEE, 2010.

\bibitem[Yadav et~al.(2023)Yadav, Choshen, Raffel, and Bansal]{yadav2023compeft}
Prateek Yadav, Leshem Choshen, Colin Raffel, and Mohit Bansal.
\newblock Compeft: Compression for communicating parameter efficient updates via sparsification and quantization.
\newblock \emph{arXiv preprint arXiv:2311.13171}, 2023.

\bibitem[Yang et~al.(2024)Yang, Yang, Hui, Zheng, Yu, Zhou, Li, Li, Liu, Huang, Dong, Wei, Lin, Tang, Wang, Yang, Tu, Zhang, Ma, Xu, Zhou, Bai, He, Lin, Dang, Lu, Chen, Yang, Li, Xue, Ni, Zhang, Wang, Peng, Men, Gao, Lin, Wang, Bai, Tan, Zhu, Li, Liu, Ge, Deng, Zhou, Ren, Zhang, Wei, Ren, Fan, Yao, Zhang, Wan, Chu, Liu, Cui, Zhang, and Fan]{qwen2}
An Yang, Baosong Yang, Binyuan Hui, Bo Zheng, Bowen Yu, Chang Zhou, Chengpeng Li, Chengyuan Li, Dayiheng Liu, Fei Huang, Guanting Dong, Haoran Wei, Huan Lin, Jialong Tang, Jialin Wang, Jian Yang, Jianhong Tu, Jianwei Zhang, Jianxin Ma, Jin Xu, Jingren Zhou, Jinze Bai, Jinzheng He, Junyang Lin, Kai Dang, Keming Lu, Keqin Chen, Kexin Yang, Mei Li, Mingfeng Xue, Na Ni, Pei Zhang, Peng Wang, Ru Peng, Rui Men, Ruize Gao, Runji Lin, Shijie Wang, Shuai Bai, Sinan Tan, Tianhang Zhu, Tianhao Li, Tianyu Liu, Wenbin Ge, Xiaodong Deng, Xiaohuan Zhou, Xingzhang Ren, Xinyu Zhang, Xipin Wei, Xuancheng Ren, Yang Fan, Yang Yao, Yichang Zhang, Yu Wan, Yunfei Chu, Yuqiong Liu, Zeyu Cui, Zhenru Zhang, and Zhihao Fan.
\newblock Qwen2 technical report.
\newblock \emph{arXiv preprint arXiv:2407.10671}, 2024.

\bibitem[Yao et~al.(2025)Yao, Hu, and Klimovic]{yao2025deltazip}
Xiaozhe Yao, Qinghao Hu, and Ana Klimovic.
\newblock Deltazip: Efficient serving of multiple full-model-tuned llms.
\newblock In \emph{Proceedings of the Twentieth European Conference on Computer Systems}, pages 110--127, 2025.

\bibitem[Yu et~al.(2024)Yu, Yu, Yu, Huang, and Li]{yu2024language}
Le Yu, Bowen Yu, Haiyang Yu, Fei Huang, and Yongbin Li.
\newblock Language models are super mario: Absorbing abilities from homologous models as a free lunch.
\newblock In \emph{Forty-first International Conference on Machine Learning}, 2024.

\bibitem[Yuval(2011)]{yuval2011reading}
Netzer Yuval.
\newblock Reading digits in natural images with unsupervised feature learning.
\newblock In \emph{Proceedings of the NIPS Workshop on Deep Learning and Unsupervised Feature Learning}, 2011.

\bibitem[Zheng and Wang(2024)]{zheng2024free}
Shenghe Zheng and Hongzhi Wang.
\newblock Free-merging: Fourier transform for model merging with lightweight experts.
\newblock \emph{arXiv preprint arXiv:2411.16815}, 2024.

\end{thebibliography}
